\documentclass{article}

\usepackage{PRIMEarxiv}

\usepackage[utf8]{inputenc} % allow utf-8 input
\usepackage[T1]{fontenc}    % use 8-bit T1 fonts
\usepackage{hyperref}       % hyperlinks
\usepackage{url}            % simple URL typesetting
\usepackage{booktabs}       % professional-quality tables
\usepackage{amsfonts}       % blackboard math symbols
\usepackage{amsmath}
\usepackage{multirow}
\usepackage{nicefrac}       % compact symbols for 1/2, etc.
\usepackage{microtype}      % microtypography
\usepackage{lipsum}
\usepackage{fancyhdr}       % header
\usepackage{graphicx}       % graphics
\graphicspath{{figures/}}     % organize your images and other figures under media/ folder

\DeclareMathOperator*{\argminB}{argmin}
\DeclareMathOperator*{\argmaxB}{argmax}

\title{Explainable Lifelong Stream Learning Based on ``Glocal'' Pairwise Fusion
}

\author{
  Chu Kiong Loo \thanks{Corresponding author} \\
  Faculty of Computer Science and Information Technology \\
  Universiti Malaya \\
  Kuala Lumpur, Malaysia \\
  \texttt{ckloo.um@um.edu.my} \\
  \And
  Wei Shiung Liew \\
  Faculty of Computer Science and Information Technology \\
  Universiti Malaya \\
  Kuala Lumpur, Malaysia \\
  \texttt{liew.wei.shiung@um.edu.my} \\
   \And
  Stefan Wermter \\
  Knowledge Technology \\
  Department of Informatics \\
  Universität Hamburg \\
  Hamburg, Germany \\
  \texttt{stefan.wermter@informatik.uni-hamburg.de} \\
}

\begin{document}
\maketitle

\begin{abstract}
Real-time on-device continual learning applications are used on mobile phones, consumer robots, and smart appliances. Such devices have limited processing and memory storage capabilities, whereas continual learning acquires data over a long period of time. By necessity, lifelong learning algorithms have to be able to operate under such constraints while delivering good performance. This study presents the Explainable Lifelong Learning (ExLL) model, which incorporates several important traits: 1) learning to learn, in a single pass, from streaming data with scarce examples and resources; 2) a self-organizing prototype-based architecture that expands as needed and clusters streaming data into separable groups by similarity and preserves data against catastrophic forgetting; 3) an interpretable architecture to convert the clusters into explainable IF-THEN rules as well as to justify model predictions in terms of what is similar and dissimilar to the inference; and 4) inferences at the global and local level using a pairwise decision fusion process to enhance the accuracy of the inference, hence ``Glocal Pairwise Fusion.'' We compare ExLL against contemporary online learning algorithms for image recognition, using OpenLoris, F-SIOL-310, and Places datasets to evaluate several continual learning scenarios for video streams, low-sample learning, ability to scale, and imbalanced data streams. The algorithms are evaluated for their performance in accuracy, number of parameters, and experiment runtime requirements. ExLL outperforms all algorithms for accuracy in the majority of the tested scenarios. 
\end{abstract}

% keywords can be removed

\section{Introduction}
In most real-world applications, data arrives continuously in real-time and is often non-repeating unless it is memorized. From this phenomenon, two paradigms are coined: continuous learning and streaming learning. Continuous learning, also known as lifelong learning \cite{thrun1995lifelong} refers to the ability to acquire knowledge continuously over a long period of time while retaining previously-learned knowledge. Streaming learning \cite{domingos2000mining} on the other hand is the ability to acquire knowledge from sequential and continuously-arriving data streams. The former encompasses machine learning techniques to adapt and reconcile old and new knowledge while minimizing loss of information and the latter prioritizes quick and efficient knowledge acquisition from high-velocity data streams. 

When developing machine learning applications for use in embedded systems such as portable digital devices, robots, autonomous vehicles, and smart appliances, not only it is necessary for the applications to have both continuous and streaming learning capabilities, but also the ability to operate in resource-limited environments. Portable devices prioritize compactness which limits how much hardware can be installed on-board the device, thus limiting its processing power, memory storage, and energy storage capabilities. Example applications include portable medical devices which use continuous learning to personalize the diagnosis based on long-term monitoring of a patient's vital signs \cite{castineira2020adding}. Personalized action recognition systems adapt to individual variances in body movements \cite{parisi2016human}. On-device learning is preferable to ensure greater customization based on the consumer's needs, as opposed to cloud-based learning where a consumer's personalized data may be considered an insignificant detail among many other consumers' data. There are several other benefits to continual on-device learning, such as decreased bandwidth requirements, better control over the consumer's privacy, and less dependence on big data. 

Conventional learning strategies minimize empirical risk by assuming a given dataset consists of independent and identically-distributed (iid) samples and shuffling them before training. In continuous learning however, this may sometimes cause catastrophic forgetting whereby learning new knowledge causes older learned knowledge to be forgotten \cite{parisi2019continual}. While there have been many research studies to address catastrophic forgetting, not all are suitable for embedded applications. 

% Need for Explainability
Recent research also stresses the need for interpretability or explainability especially for machine learning algorithms used in critical applications that directly affect human well-being. The main criterion of an explainable learning model is being able to show its thought-processes step-by-step from the input to the final decision, improving human trust in the system \cite{nomura2011relationships} \cite{biran2017explanation} and debug potentially problematic decisions \cite{kulesza2015principles}. The current generation of continual learning systems lacks the ability to self-diagnose their decisions. A common problem involving self-supervised or unsupervised learning systems is when the data stream consists of undetected bias or garbage data, which would negatively impact the model. By implementing explainability in continual learning models, it would be possible to debug the learning process and identify problematic data before use. As of the time of writing this paper, state-of-the-art continual learning architectures such as Streaming Linear Discriminant Analysis (SLDA) \cite{pang2005incremental} did not have explainability capabilities while explainable learning architectures such as the eXplainable Deep Neural Networks (xDNN) \cite{angelov2020towards} have been tested with several continuous learning scenarios but not under streaming learning conditions \cite{hayes2022online}. 

% explain slda weaknesses and improvements?

We argue the need for the following capabilities in streaming, explainable, continually learning architectures:
\begin{enumerate}
    \item Learn from a continuous data stream in a single pass in environments where computational resources and data storage is highly constrained.
    \item Acquire knowledge from data in any order while maintaining resilience against loss of previously learned information.
    \item Learn efficiently and generalize well with minimal labeled examples.
    \item Explain model decisions at the intermediate and final stages of the decision-making process.
\end{enumerate}

This study investigates explainable continual and streaming learning specifically for embedded devices. The paper presents several research contributions in this field.
\begin{enumerate}
    \item We propose a modified SLDA architecture to utilize a prototype-based architecture to address the issues of catastrophic forgetting and the stability-plasticity dilemma, the balancing between the network's ability to retain and integrate knowledge. 
    
    \item We introduce a collective inference strategy to enhance classification accuracy by combining inferences from two levels: local inferences at the prototype level (i.e. "Among the examples in Class A, which example is the closest match to the input?") and global inferences at the class level (i.e. "Among all the classes, which is the closest match to the input?"). 
    
    \item We formulate an explainable lifelong stream learning model with single-pass learning.
    
    \item We conduct a series of benchmark tests and observe how the proposed model performed relative to other continual learning models using established datasets and continual learning scenarios.
\end{enumerate}

\section{Problem Definition}\label{sec2}
% The challenge of integrating XAI to lifelong learning
% Existing XAI majorly based on posthoc methods, which is not applicable to lifelong learning 

Integrating explainability with online continual learning applications is challenging due to a number of factors. In online continual learning, examples are only presented once and may not be repeated unless they are stored in memory. Online continual learning models receive limited data and have a short time to learn from them. Given the scarcity of training data and learning time, it is difficult for most explainable algorithms to accurately model the concept enough to generate adequate explanations \cite{vinyals2016matching}.  

In addition, the data obtained from the continuous learning process is constantly changing. This means that the information learned by the models may also change over time and lose relevance. While deep learning models are capable of achieving high accuracy, their opaque nature makes it a challenge to generate explanations for their predictions \cite{rai2020explainable} \cite{ribeiro2016should}.

Another major challenge is implementing online continual learning algorithms on embedded devices with limited memory capacity and processing power. This restriction makes it difficult to deploy complex algorithms or algorithms that take up a lot of storage space \cite{lane2017squeezing} \cite{olsson2018challenges}. 

Models that provide understandable explanations typically compromise on the accuracy of the results. Balancing between explainability and accuracy is a challenge when employing explainable methods in online continual learning applications \cite{dziugaite2020enforcing}.

\section{Related Work}
% Focus on stream learning review
% On-device learning
% Lifelong learning

\subsection{Streaming Learning}

In streaming learning scenarios, machine learning algorithms are required to learn from a continuous stream of non-repeating training samples in a single pass. The algorithms must also be capable of being evaluated at any point during the stream and prior training samples are not stored for retraining. In real-life applications, contextual information may not always be available. Several prototype-based classifiers such as ARTMAPs \cite{carpenter1991artmap} were developed to learn from non-stationary data. However, the presentation order of training data significantly affects the performance of ARTMAPs. Various methods were developed to optimize ARTMAP performance \cite{palaniappan2009using, yaghini2013gofam, liew2015affect} but they were computationally intensive and therefore unsuitable for real-time applications. 

Streaming Linear Discriminant Analysis (SLDA) \cite{pang2005incremental} extends the conventional Linear Discriminant Analysis (LDA) architecture to support incremental learning from data streams. SLDA stores a running mean for each unique class and a shared covariance matrix. During inference, SLDA classifies a given input to the most likely class using the class means and covariance matrix. The softmax methods used with conventional neural networks are analogous to the LDA's estimated posterior distribution \cite{lee2018simple}. Deep-SLDA \cite{hayes2020lifelong} pairs SLDA with a convolutional neural network (CNN) acting as a feature extractor for high-dimensional inputs such as images. The performance of the model surpasses that of state-of-the-art streaming learning and incremental batch learning algorithms.

\begin{figure*}
    \centering
    \includegraphics[width=0.99\linewidth]{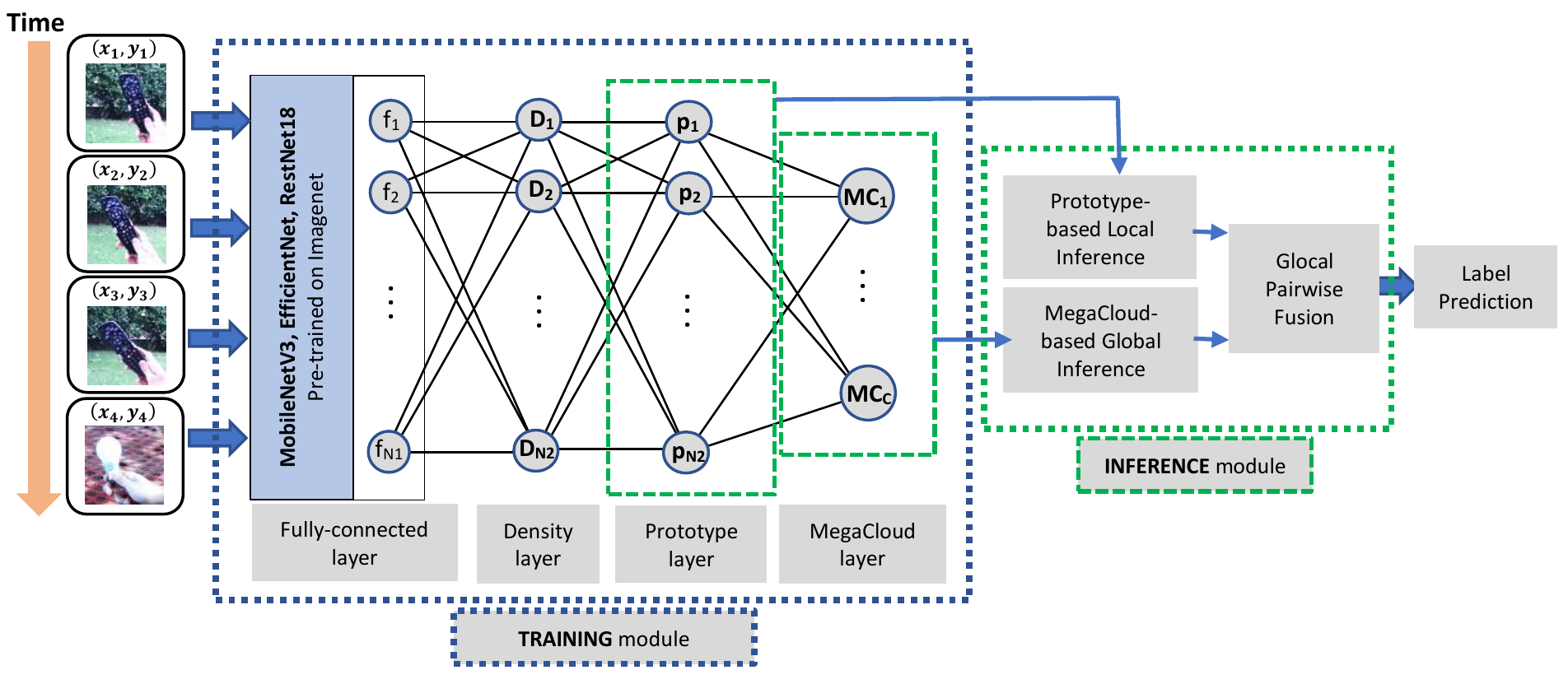}
    \caption{Explainable Lifelong Learning architecture. Training images produce feature vectors in the fully-connected layer of a pre-trained CNN. Density of the feature vectors are then computed to determine if the training images should be assigned to an existing prototype or to initialize a new prototype. All prototypes belonging to one class label are assigned to one MegaCloud. Inference is performed once at the local level, another at the global level. Local inference matches the inferenced image to the most similar prototype while global inference matches the image to the most similar MegaCloud. Both inference decisions are then combined using a glocal pairwise fusion matrix to obtain the final model decision. }
    \label{fig_architecture}
\end{figure*}

\subsection{Continual Embedded / On-Device Learning}

Although streaming learning algorithms have been developed to reduce catastrophic forgetting, they don't meet certain requirements for embedded applications. Disqualifying criteria include the high storage and computation requirements of batch learning techniques, and needing task labels during inference \cite{hayes2020lifelong} \cite{hayes2020remind}. Another requirement for continual embedded learning is the ability to generalize from a very small number of training samples. Algorithms with this capability are commonly known as ``low-shot'' continual learning algorithms \cite{ayub2020cognitively, ayub2021fsiol, tao2020few, tao2020topology}.

Several CNNs were made to meet the need for on-device learning, balancing accuracy of classification with speed of processing. Networks with efficient computation and reduced memory requirements include MobileNet \cite{howard2017efficient}, SqueezeNet \cite{iandola2016squeezenet}, ShuffleNet \cite{zhang2018shufflenet} and CondenseNet \cite{huang2018condensenet}. Other methods to reduce memory requirements include deep network pruning \cite{alvarez2016learning, hu2016network, li2016pruning, louizos2017learning}, quantization \cite{courbariaux2016binarized, kim2016bitwise, wu2016quantized, zhou2016dorefa, zhou2017incremental, jacob2018quantization, krishnamoorthi2018quantizing}, and model compression or network distillation \cite{buciluă2006model} \cite{hinton2015distilling}.

A comprehensive study was performed to compare several continual learning algorithms and CNNs as feature extractors in multiple scenarios \cite{hayes2022online}. The models were tested on their robustness to scale, on imbalanced class distribution, and on temporally correlated video streams. The models were then evaluated on the basis of classification accuracy, number of parameters, and experiment runtime. We use the same experiment protocols to evaluate the performance of our proposed continual learning model against other algorithms.

\subsection{Explainable Prototype-Based Learning Models}

The architecture of CNNs is designed to maximize predictive accuracy through a series of convolutional steps. CNNs are considered ``black box'' models due to how difficult it is to explain how they arrive at a specific classification decision for a given input. CNNs are typically interpreted post hoc: the model's decisions are obtained first before backtracking and generating justifications \cite{li2018deep}. A popular explainable technique uses class activation mappings (CAMs) and gradient-weighted CAMs (Grad-CAMs) \cite{zhou2016learning} \cite{selvaraju2017grad} to highlight discriminative features on input images. Such post hoc interpretability techniques are usually approximations as opposed to in-depth explanations of the cause-and-effect relations and reasoning. 

Prototype-based classifiers such as ARTMAPs \cite{carpenter1991artmap} and self-organizing networks \cite{parisi2018lifelong} group training samples according to their proximity in the feature space \cite{biehl2016prototype}. Each group or cluster of training samples can be represented by the closest centroid or prototype \cite{liu2018memory}. % The similarity between an observed sample and the prototype centroids is measured in a latent space, created using autoencoders or CNNs \cite{li2018deep}, or by using Euclidean distance computation expressed in terms of convolutional operations and dot product operations for efficient computation \cite{biehl2014distance} \cite{nebel2017types}. 

xDNN \cite{angelov2020towards} is a prototype-based classifier with the ability to generate explanations for deep neural networks. The prototypes in the architecture are used to generate linguistic IF-THEN rules. xDNN employs empirically derived probability distribution functions based on local densities and global multivariate generative distributions \cite{angelov2019empirical}. The prototype-based architecture and algorithm are suitable for transfer learning and continuous learning without retraining. xDNN outperforms state-of-the-art approaches in accuracy and computational simplicity in benchmark tests \cite{angelov2020towards} \cite{angelov2018deep}. To summarize, xDNN is an explainable feed-forward neural network with an incremental learning algorithm adding new prototypes to reflect the dynamic data stream \cite{soares2019novelty}.

\section{Methodology}
% Elaborate from xDNN papers

The proposed Explainable Lifelong Learning (ExLL) model is a feed-forward neural network with an incremental learning algorithm and a self-organizing topology. Inputs to the network are typically images passed through a convolutional neural network to extract both abstract and discriminative features from the fully-connected layer. The architecture of ExLL enhances the functionality of the xDNN \cite{angelov2020towards} with a few modifications for implementing a variant of SLDA \cite{pang2005incremental}, namely MegaCloud-based global inference and prototype-based local inference. A pairwise fusion method \cite{ko2007pairwise} is used to combine the global and local inferences into a ``glocal'' inference. 

% divide into training and inference subsections
\subsection{Training the Explainable Continual Learning Model}
% Towards explainable deep neural networks (xDNN)

Figure \ref{fig_architecture} shows the ExLL model's layers. CNN weights are pre-trained with image datasets such as ImageNet. Images are passed through the CNN, and the activations of the last hidden fully-connected layer in the CNN are taken as the discriminative feature vectors to be learned by the ExLL. 

Similar to xDNN, the two main components of the proposed model are the Prototypes Layer and the MegaClouds Layer \cite{angelov2020towards}. Input feature vectors are represented as data points in a multi-dimensional topology. Data points that are close to each other can be considered as a ``data cloud'' encompassing an area of influence between the data points. A data cloud can be represented by a composite feature vector known as a centroid or ``prototype'', calculated as the average of all points in the data cloud. A prototype is typically assigned a class label based on the labels of the majority points in the data cloud. Each prototype is independent and distinct from each other, representing the local peaks of the data distribution sharing the same class label. 

Furthermore, adjacent data clouds sharing the same class label can be grouped into a larger structure known as a ``MegaCloud''. Where a prototype represents an instance-based prototype of a class label, a MegaCloud is a category-based prototype of the class label. 

Training the ExLL takes place as follows:

\begin{enumerate}
    \item An image $I_i$ at time stamp $i$ and belonging to class $k$ is passed through a CNN. The subsequent feature vector $\Tilde{x}_i$ is obtained from the fully-connected layer, and then normalized:
    \begin{equation}
        x_{i} = \frac{\Tilde{x}_i}{\Vert \Tilde{x}_i \Vert}
    \end{equation}
    where $\Vert \cdot \Vert$ is the vector norm. 

    \item The ExLL’s global meta-parameters are updated:
    \begin{equation}
        \begin{split}
            \hat{\mu}_i & = \frac{i-1}{i} \hat{\mu}_{i-1} + \frac{1}{i} x_i \\
            \hat{\mu}_1 & = x_1
        \end{split}
    \end{equation}
    \begin{equation}
        \begin{split}
            \hat{\sigma}_i & = \frac{i-1}{i} \hat{\sigma}_{i-1} + \frac{1}{i} \Vert x_i \Vert^2 \\
            \hat{\sigma}_1 & = \Vert x_1 \Vert^2 = 1
        \end{split}
    \end{equation}
    \begin{equation} \label{eq_density_update_cov}
        \begin{split}
            \hat{\xi}_i & = \frac{i-1}{i} \hat{\xi}_{i-1} + \frac{1}{i} (x_i - \hat{\mu }_i) (x_i - \hat{\mu }_i)^T \\
            \hat{\xi}_1 & = (x_1) (x_1)^T
        \end{split}
    \end{equation}
    where $\hat{\mu}$ is the global average of all training samples, $\hat{\sigma}$ is the global variance, and $\hat{\xi}$ is the inter-class global covariance matrix.

    \item While the global meta-parameters represent the cross-class topology of the ExLL, local meta-parameters represent the within-class topologies for each class. If $x_i$ has a novel class, the number of unique class labels is incremented, $k \leftarrow k + 1$, and the local meta-parameters for the new class $k$ are initialized as follows:
    \begin{equation}
        \begin{split}
            i_k \leftarrow 1 \\
            g_k \leftarrow 1 \\
            \mu_{k, 1} \leftarrow x_i \\
            \sigma_{k, 1} \leftarrow \Vert x_i \Vert^2 \\
            % \xi_{k, 1} \leftarrow (x_i) (x_i)^T \\
            E_{k, 1, 1} = 0 \\
        \end{split}
    \end{equation}
    Here $i_k$ denotes the number of inputs where class $k$ was observed during training, $g_k$ counts the prototypes in class $k$, $\mu_{k, 1}$ is the class mean, $\sigma_{k, 1}$ is the class scalar product, and $E_{k, 1, 1}$ is a topological map of edge connections between within-class prototypes.

    Additionally, the novel class is used to initialize the first prototype of a new MegaCloud:
    \begin{equation}
        \begin{split}
            p_{k, 1} \leftarrow x_i \\ 
            \text{S}_{k, 1} \leftarrow 1 \\
            r_{k, 1} \leftarrow r^* \\
            \hat{I}_{k, 1} \leftarrow I_i \\
        \end{split}
    \end{equation}
    where $p_{k, 1}$ is the first prototype for class $k$, $\text{S}_{k, 1}$ is the number of training samples associated with the prototype, $r_{k, 1}$ is the prototype's radius of influence initialized to a default value $r^* = \sqrt{2 - 2 \cos{30^{\circ}}}$ \cite{angelov2019empirical}, and $\hat{I}_{k, 1}$ keeps a record of all input images associated with this prototype, without actually storing the images themselves. The network then waits for the next input.

    However, if the input presents a known class, then the prototype layer is updated in response. The local meta-parameters are updated similar to the global meta-parameters as follows:
    \begin{equation}
        \mu_{k, i_k} = \frac{i_k - 1}{i_k} \mu_{i_k-1} + \frac{1}{i_k} x_i \\
    \end{equation}
    \begin{equation}
        \sigma_{k, i_k} = \frac{i_k - 1}{i_k} \sigma_{i_k - 1} + \frac{1}{i_k} \Vert x_i \Vert^2 \\
    \end{equation}
    Class $k$'s mean $\mu_{k, i_k}$ and scalar product $\sigma_{k, i_k}$ are updated online. The input $x_i$ is then passed to the density layer.
    
    \item \textbf{Density layer}. This layer defines the mutual proximity of the training images relative to the data space defined by the feature vectors. The density of input $x_i$ relative to class $k$, $D(k, x_i)$ can be computed online \cite{angelov2012autonomous}:
    
    \begin{equation}
        D(k, x_i) = \frac{1}{1 + \Vert x_i - \mu_{k, i_k} \Vert^2 + \sigma_{k, i_k} - \Vert \mu_{k, i_k} \Vert^2}
    \end{equation}
    
    % \begin{equation} \label{eq_density_update_mean}
    %     \begin{split}
    %         \mu_{\hat{S}} & = \frac{\hat{S}-1}{\hat{S}} \mu_{\hat{S}-1} + \frac{1}{\hat{S}} x_i \\
    %     \end{split}
    % \end{equation}
    % \begin{equation} \label{eq_density_update_scalar}
    %     \begin{split}
    %         \sigma_{\hat{S}} & = \frac{\hat{S}-1}{\hat{S}} \sigma_{\hat{S}-1} + \frac{1}{\hat{S}} \Vert x_i \Vert^2 \\
    %     \end{split}
    % \end{equation}
    % where $\hat{S} = \Sigma_{j=1}^{P_k} S_{j}$ is the total number of training samples for class $k$ that have been presented.
    
    % \item \textbf{Typicality} layer. Typicality $\tau$ is an empirically derived form of probability distribution function:
    % \begin{equation}
    %     \tau(x_i) = \frac{\sum^{C}_{k=1} \sum^{g_k}_{j=1} S_{k, g_k} D(k, x_i)}{\sum^{C}_{k=1} \sum^{g_k}_{j=1} S_{k, g_k} \int_{-\infty}^{\infty} D(k, x_i) dx}
    % \end{equation}
    % where the integral $\int_{-\infty}^{\infty} \tau dx = 1$, $S_{k, g_k}$ is the support, or number of training samples associated with the $g_k$-th prototype of class $k$, and $C$ is the total number of unique classes.

    \item \textbf{Prototype} layer. When an input $x_i$ is presented, the nearest and second-nearest within-class prototypes, $b_1$ and $b_2$, are identified using Mahalanobis distance \cite{mclachlan1999mahalanobis}:
    \begin{equation} \label{eq_winning_node1}
        b_1 = \argminB_{j=1,...,g_k} \frac{(x_i - p_j)^T (x_i - p_j)}{\hat{\xi}}
    \end{equation}
    \begin{equation} \label{eq_winning_node2}
        b_2 = \argminB_{j=1,...,g_k; j \neq b1} \frac{(x_i - p_j)^T (x_i - p_j)}{\hat{\xi}}
    \end{equation}

    A density condition then tests if $x_i$ is inside the distribution of existing prototypes:
    \begin{equation} \label{eqn_density}
        \begin{split}
            \text{IF } & D(k, x_i) > \max_{j=1,...,g_k} D(k, p_j) \\
            \text{OR } & D(k, x_i) < \min_{j=1,...,g_k} D(k, p_j) \\
            \text{THEN } & \text{add a new data cloud } (g_k \leftarrow g_k + 1)
        \end{split}
    \end{equation}
    
    If Condition \ref{eqn_density} is met, then the input $x_i$ is considered outside the influence radius of the current prototypes and is sufficiently novel. $x_i$ is then used to initialize a new data cloud:
    \begin{equation} \label{eqn_prototype_new} 
        \begin{split}
            g_k \leftarrow g_k + 1 \\ 
            \text{S}_{k, g_k} \leftarrow 1 \\
            p_{k, g_k} \leftarrow x_i \\ 
            r_{k, g_k} \leftarrow r_0 \\
            \hat{I}_{k, g_k} \leftarrow I_i \\
            E_{k, g_k, b_1} \leftarrow 1; E_{k, b_1, g_k} \leftarrow 1; \\
        \end{split}
    \end{equation}
    
    Otherwise, if Condition \ref{eqn_density} is not met, the parameters are updated for the closest matching prototype $b_1$:
    \begin{equation} \label{eqn_prototype_update} 
        \begin{split}
            \text{S}_{k, b_1} \leftarrow \text{S}_{k, b_1} + 1; \\
            p_{k, b_1} \leftarrow \frac{\text{S}_{k, b_1}-1}{\text{S}_{k, b_1}}p_{k, b_1} + \frac{1}{\text{S}_{k, b_1}}x_i; \\
            r_{k, b_1} \leftarrow \sqrt{\frac{r^2_{k, b_1} + (1 - \Vert p_{k, b_1} \Vert^2)}{2}} \\
            \hat{I}_{k, b_1} \leftarrow \hat{I}_{k, b_1} + I_i \\
            E_{k, b_1, b_2} \leftarrow E_{k, b_1, b_2} + 1 \\
            E_{k, b_2, b_1} \leftarrow E_{k, b_2, b_1} + 1
        \end{split}
    \end{equation}
    
    $E_{k}$ is a square matrix sized $g_k$ for encoding the edges between local prototypes. Whenever a training sample activates the closest prototype $b_1$ and second-closest prototype $b_2$, $E_{k, b_1, b_2} $ and $E_{k, b_2, b_1}$ are incremented. The map $E_{k}$ can then be used for visual evaluation of the spatial relationship between prototypes or for encoding frame of reference transformations \cite{weber2007self} with each prototype representing one frame of reference.
    
    The prototype layer is the basis of local explainability of the ExLL model. Each prototype is an independent and distinct centroid shaped by associated training inputs. As each prototype records the associated training images in $\hat{I}$, a set of linguistic IF-THEN rules are formulated as:
    \begin{equation}
        \begin{split}
        R_c: \text{IF } (I \mathtt{\sim} \hat{I}_{k, 1}) \text{ OR } ... \text{ OR } (I \mathtt{\sim} \hat{I}_{k, g_k}) \\
        \text{ THEN } (\text{class is \textit{k}})
        \end{split}
    \end{equation}
    where $\mathtt{\sim}$ indicates similarity or fuzzy degree of membership to a prototype, $k=\{1,...,C\}$ is the class, and $I_i$ denotes an input image.

    % original XDNN mixes prototypes in one topology; current method splits topology by class
    \item \textbf{MegaClouds} layer. This layer is the basis of global explainability of the ExLL model. Each MegaCloud is used to facilitate explainability at the class level. Explainable rules generated from MegaClouds have the following format:
    \begin{equation}
        R_k : \text{IF }(x_i \mathtt{\sim} \text{MC}_k) \text{ THEN }(\text{class is \textit{k}})
    \end{equation}
    where $\text{MC}_k$ is the MegaCloud for the class $k$.
\end{enumerate}

\begin{figure*}
    \centering
    \includegraphics[width=0.99\linewidth]{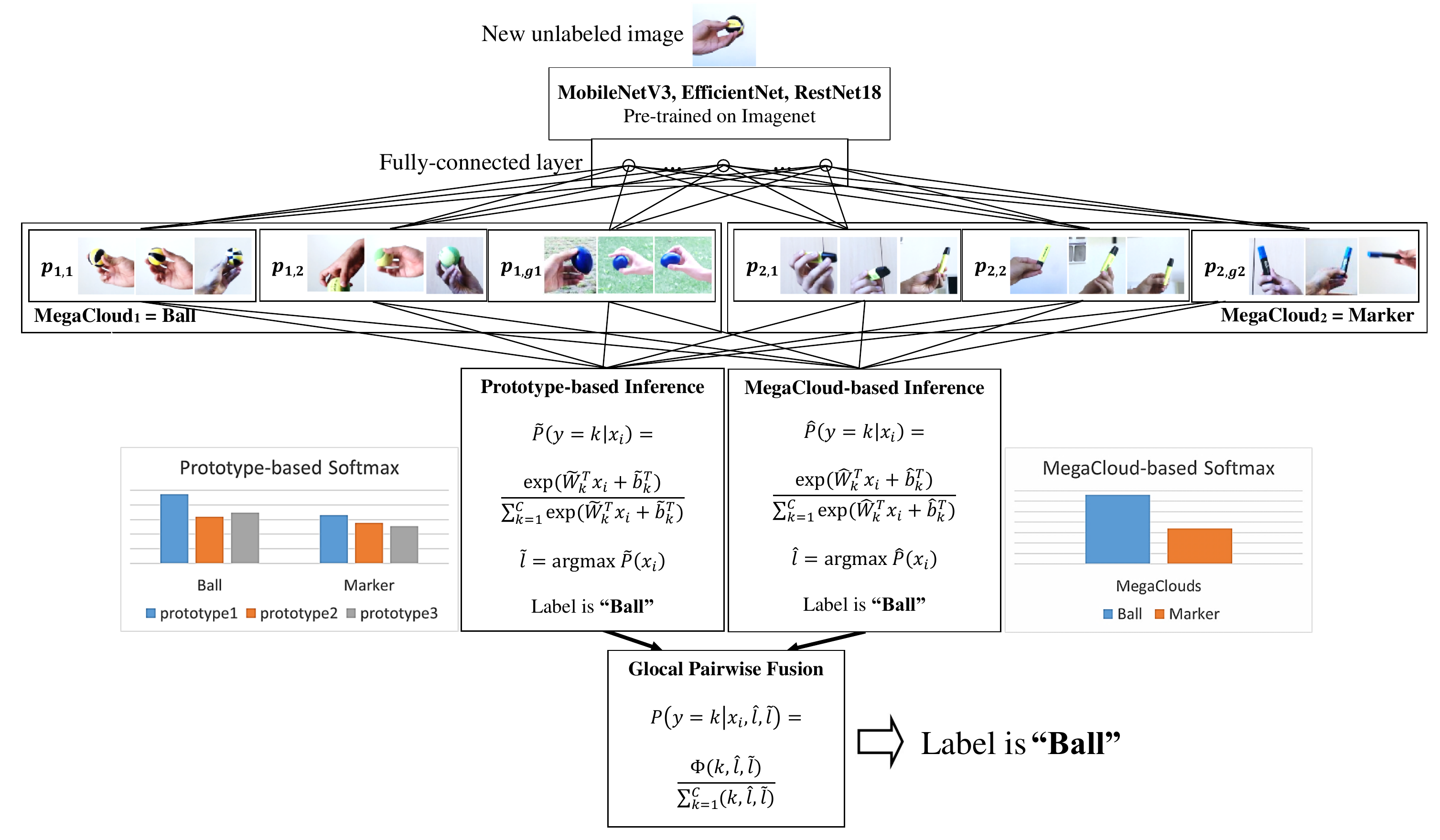}
    \caption{The process of image inference for the proposed ExLL.}
    \label{fig_inference}
\end{figure*}

\subsection{Inferring the Explainable Continual Learning Model}

Figure \ref{fig_inference} illustrates the process where a given image is inferred. Prototype-based inference (PrInf) considers the local discriminative ability between individual prototypes while MegaCloud-based inference (McInf) globally discriminates between classes. Both types of inference have their strengths and weaknesses which adapt depending on the class distribution of the used dataset. 

Pairwise fusion (PF) is used as a method for combining local and global inferences to achieve better performance than either technique alone, hence the term ``glocal pairwise fusion''. A PF matrix encodes the association between PrInf and McInf during training without prior knowledge of the performance of either technique or the distribution of the dataset.

\begin{enumerate}
    \item Shrinkage regularization is used to compute the precision matrix from the covariance matrix $\hat{\xi}$:
    \begin{equation}
        \Lambda = [(1-\epsilon) \hat{\xi} + (\epsilon) I]^{-1}
    \end{equation}
    where $I$ is an identity matrix and $\epsilon=1e^{-4}$ regulates shrinkage. 

    \item \textbf{Prototype-based inference} assembles the prototypes of a class $k$, $\tilde{p}_k = \{p_{1, 1}, ..., p_{k, g_k}\}$, and $\Lambda$ to construct local weights $\tilde{W}_k$ and local bias $\tilde{b}_k$: 
    \begin{equation}
        \begin{split}
            \tilde{W}_k & = \Lambda \tilde{p}_k \\
            \tilde{b}_k & = -\frac{1}{2} (\tilde{p}_k \cdot \tilde{W}_k)
        \end{split}
    \end{equation}
    
    Subsequently, the posterior distribution $\tilde{P}$ and label prediction $\tilde{l}$ are formulated as follows:
    \begin{equation} \label{eq_prinf}
        \begin{split}
            \tilde{P} (y=k | x_i) & = \frac{\exp (\tilde{W}^T_k x_i + \tilde{b}^T_k)}{\Sigma_{k=1}^{C} \exp (\tilde{W}^T_k x_i + \tilde{b}^T_k)} \\
            \tilde{l} & = \argmaxB_{k=1,...,C} \tilde{P}(y=k | x_i)
        \end{split}
    \end{equation}

    \item \textbf{MegaCloud-based inference} assembles all class mean vectors $\hat{\mu} = \{\mu_1, ..., \mu_C\}$, and $\Lambda$ to construct global weights $\hat{W}$ and bias $\hat{b}$: 
    \begin{equation} 
        \begin{split}
            \hat{W} & = \Lambda \hat{\mu} \\
            \hat{b} & = -\frac{1}{2} (\mu \cdot \hat{W})
        \end{split}
    \end{equation}
    and the subsequent posterior distribution $\hat{P}$ and label prediction $\hat{l}$ are formulated as follows:
    \begin{equation} \label{eq_mcinf}
        \begin{split}
            \hat{P}(y=k | x_i) & = \frac{\exp (\hat{W}^T_k x_i + \hat{b}^T_k)}{\Sigma_{k=1}^{C} \exp (\hat{W}^T_k x_i + \hat{b}^T_k)} \\
            \hat{l} & = \argmaxB_{k=1,...,C} \hat{P}(y=k | x_i)
        \end{split}
    \end{equation}

    \item \textbf{Glocal pairwise fusion} \cite{ko2007pairwise} is used for combining the two inferences. During training, given an input vector's class $k_i$, the local class prediction $\tilde{l}$, and the global class prediction $\hat{l}$, pairwise fusion encodes the relationship as:
    \begin{equation}
        \Phi(k_i, \hat{l}, \tilde{l}) \leftarrow \Phi(k_i, \hat{l}, \tilde{l}) + 1
    \end{equation}
    where $\Phi$ is a 3-dimensional matrix encoding the cumulative interactions between the actual label and the predicted labels from the local PrInf predictions and from the global McInf predictions. $\Phi$ is updated online as additional training inputs are presented. Other rules for updating the matrix, such as using confidence-based increments \cite{ko2007pairwise}, can be applied instead of simplified increments. 

    When performing inference on an object $x_i$ with an unknown class, global inference $\hat{l}$, local inference $\tilde{l}$, and $\Phi$ are used for estimating the glocal class probabilities $P(y=k | x_i, \hat{l}, \tilde{l})$ and glocal label prediction $L$: 
    \begin{equation} \label{fusion_probability}
        \begin{split}
            P(y=k | x_i, \hat{l}, \tilde{l}) & = \frac{\Phi(k, \hat{l}, \tilde{l})}{\Sigma_{k=1}^{C} \Phi(k, \hat{l}, \tilde{l})} \\
            L & = \argmaxB_{k=1,...,C} (P(y=k | x_i, \hat{l}, \tilde{l}))
        \end{split}
    \end{equation}
\end{enumerate}

\begin{figure}
    \centering
    \includegraphics[width=0.75\linewidth]{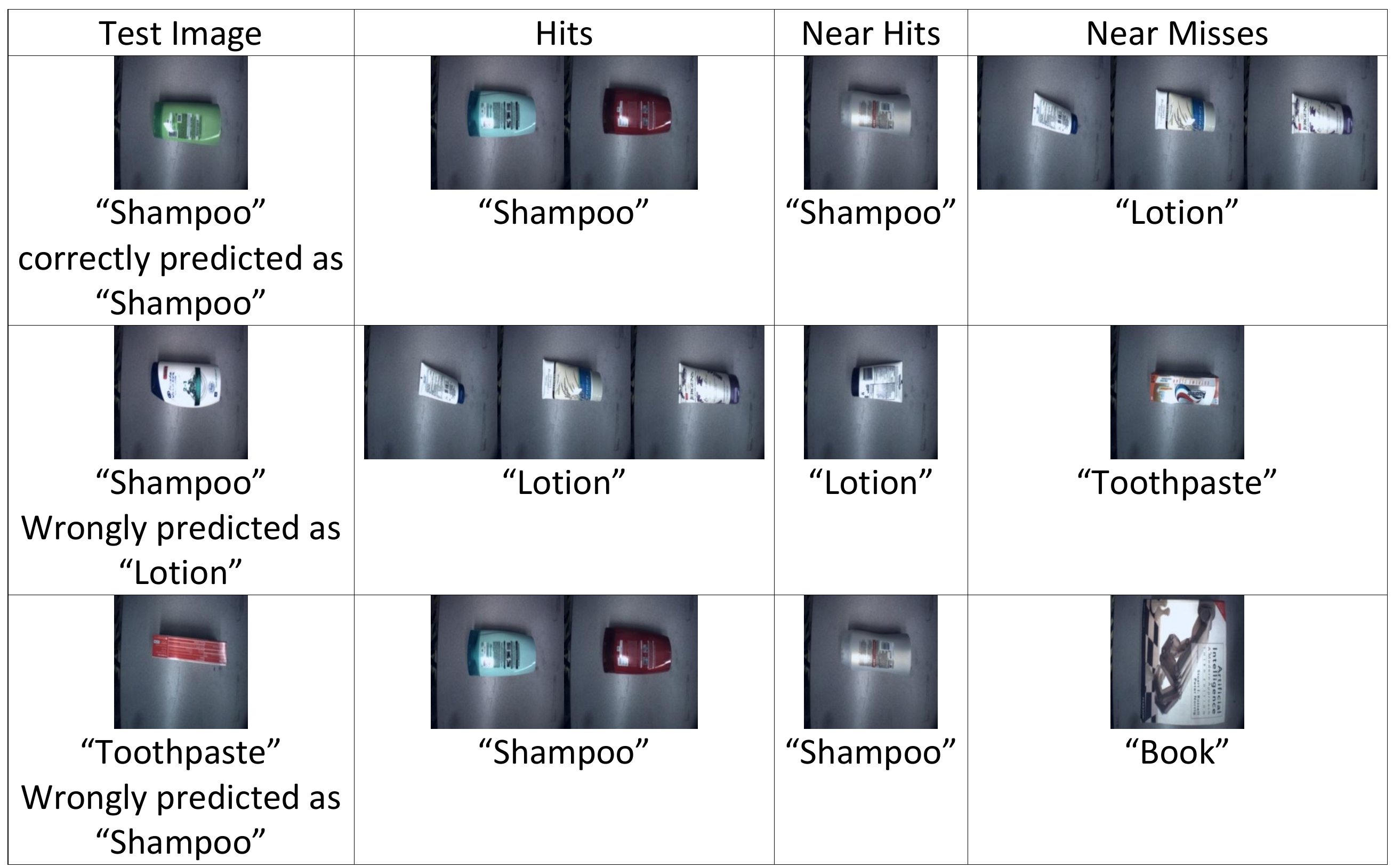}
    \caption{Example of ``Hits'', ``Near Hits'', and ``Near Misses'' for the F-SIOL-310 dataset. The test image in the top row is an example of a True Positive result, the test image in the middle row is a False Negative result, and the test image in the bottom row is a False Positive result.}
    \label{fig_nearhitnearmiss}
\end{figure}

\begin{figure}
    \centering
    \includegraphics[width=0.75\linewidth]{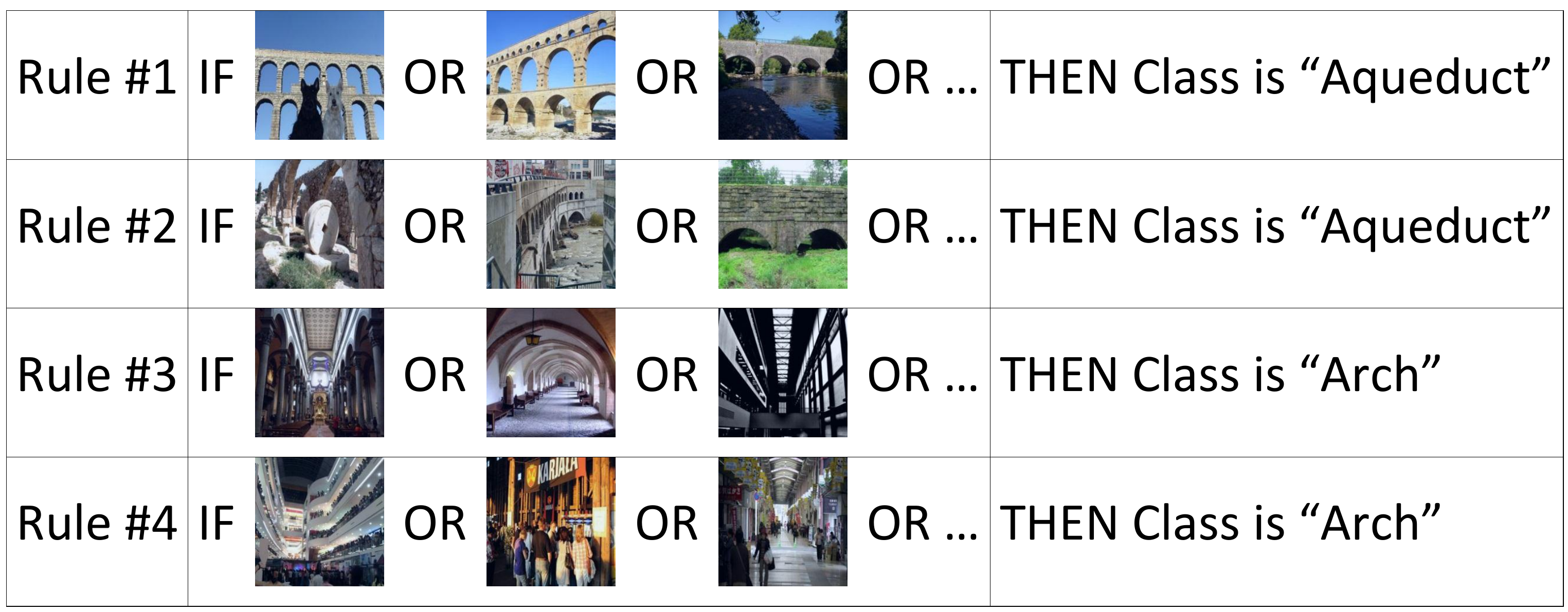}
    \caption{Explainable rules extracted from prototypes for the classes ``Aqueduct'' and ``Arch'' from Places-365. Each rule is made up of training images associated with the corresponding prototype.}
    \label{fig_extractedrules}
\end{figure}

\subsection{Explainability: Inference and Rule Generation}

ExLL incorporates the element of explainability at the inference stage, so that label predictions can be explained in terms of ``Hits'', ``Near Hits'' and ``Near Misses'' \cite{herchenbach2022explaining}. Given an image with a known label $k$ to be classified, Equation \ref{fusion_probability} produces the best-matching label $L_1$ and second-best matching label $L_2$. Going back to Equations \ref{eq_winning_node1} and \ref{eq_winning_node2}, the predicted class labels $L_1$ and $L_2$ each have a best-matching prototype ($b_{L1,1}$ and $b_{L2,1}$) as well as a second-best matching prototype ($b_{L1,2}$ and $b_{L2,2}$).

As explained in Equations \ref{eqn_prototype_new} and \ref{eqn_prototype_update}, each prototype $g_k$ is updated with a record of all associated training images: $\hat{I}_{g_k}$. When the winning prototype $b_{L1,1}$ is selected for the winning label $L_1$ during inference, $\hat{I}_{b_{L1,1}}$ is referenced to retrieve the images used to train the prototype. The retrieved images are then shown as a visual explanation, i.e. ``Hit'', as to why the inferenced image is assigned to the best-matching prototype. Where the best-matching prototype is selected based on spatial proximity, laymen can observe the retrieved images for visual comparison against the inferenced image. 

A similar comparison is made, ``Near Hit'', by showing the training images associated with the second-best matching prototype, $\hat{I}_{b_{L1,2}}$. Lastly, ``Near Miss'' shows the training images associated with the winning prototype $b_{L2,1}$ for the next-best label $L_2$: $\hat{I}_{b_{L2,1}}$. The visual explanations provided by the ``Near Hits'' and ``Near Misses'' describe the decision boundary of the ExLL's prediction. In edge cases where the predictions are ambiguous, the visual comparison of ``Hits'', ``Near Hits'', and ``Near Misses'' informs the user of possible alternatives.

Figure \ref{fig_nearhitnearmiss} demonstrates an example of a correct prediction and two wrong predictions. The top row illustrates the explanation for a True Positive prediction. An image of a shampoo bottle is correctly classified and the training images shown under ``Hits'' justify the selection of the best-matching prototype due to their visual similarity. The training image from the second-best matching prototype, shown under ``Near Hits'', also explains why the prototype is not selected due to the visual dissimilarity to the inferenced image. Lastly, ``Near Misses'' show why the test image is almost incorrectly classified as ``Lotion'' by showing the associated training images of the best prototype from the next-best class label. Given an incorrect prediction such as the False Negative result in the middle row and the False Positive result in the bottom row, the training images shown for ``Hits'', ``Near Hits'', and ``Near Misses'' explain why the ExLL made the mistake. 

The records $\hat{I}_{g_k}$ are used for visualizing explainable rules. One rule is generated from one prototype. The visualization of explainable rules reveal hidden information in each clustered prototype, as shown in Figure \ref{fig_extractedrules}. For example, the prototype associated with Rule 1 consisted of aqueducts with clear blue skies in the background. In comparison, the training images associated with the prototype for Rule 2 do not have a visible background. Similarly, the prototype used for generating Rule 3 consists of images of arches over long hallways while the prototype for Rule 4 mainly contains images of arches with people in it. This information is not immediately visible to users since the images have been converted into feature vectors, but can be shown when the images are retrieved after training the model.

\section{Experiment Setup}

% Backborn Archietecture
% Online continual learning models (Benchmark comparison)
% Datasets (FSIOL, OpenOris, Place 365, Place-LT)
% Experimental protocol - data ordering study, etc.
% Performance metrics 

Given a continuous stream of images where $X_t$ is an image at time $t$, a neural network classifier $F(\cdot)$ is trained incrementally using supervised online continual learning, producing a predicted label $\hat{y_t} = F(G(X_t))$. The backbone CNN $G(\cdot)$ is pre-trained on large image datasets such as the ImageNet-1k dataset \cite{russakovsky2015imagenet}. Subsequently, feature vectors are obtained from the last hidden fully-connected layer in response to training images fed to $G(\cdot)$, which are then passed to $F(\cdot)$ for learning. The intermediate layers in $G(\cdot)$ are frozen after pre-training to prevent knowledge drift, i.e. the learned representations in $F(\cdot)$ are no longer up-to-date. 

With this configuration, eight online continual learning strategies were studied for $F(\cdot)$ and five backbone architectures were studied for $G(\cdot)$. These studies are detailed in the following subsections.

\subsection{Backbone Architectures}

Three backbone CNN architectures were selected for comparison for their compact size, effectiveness, and classification accuracy when trained and tested on the ImageNet dataset. 

\textbf{MobileNet-v3} \cite{howard2019searching} is the successor of two previous architectures created for mobile and embedded applications \cite{howard2017efficient} \cite{sandler2018inverted}. The CNN incorporates several strategies for efficient and accurate inference under real-time and resource-constrained scenarios. Depth-wise separable convolutions are used in conjunction with linear bottleneck layers to reduce computational cost without negatively impacting performance. Two versions are compared in this study. MobileNet-v3 Small (\textbf{MNet-S}) is more efficient but displays worse classification performance compared to MobileNet-v3 Large (\textbf{MNet-L}) which is more resource-intensive but shows better classification performance. 

\textbf{EfficientNet} \cite{tan2019efficientnet} utilizes neural architecture search (NAS) to automate the selection of an optimal architecture to achieve a good tradeoff between performance and model complexity. Like MobileNet-v3, EfficientNet utilizes depth-wise separable convolutions and linear bottleneck layers to reduce computational cost, making it suitable for usage in embedded and mobile applications with limited computing resources. EfficientNet refers to a family of models with varying complexity. Two models with the least complexity are compared in this study: EfficientNet-B0 (\textbf{ENet-B0}) and EfficientNet-B1 (\textbf{ENet-B1}).

\textbf{ResNet} \cite{he2016deep} makes use of residual blocks allowing information to skip one or more convolutional layers and is efficient when involving very deep networks. During training, residual representations measure the differences between the actual output from each block and the desired output. Learning is performed by updating the convolutional weights to make the residual representations more accurate. ResNet includes several types of models with varying complexity. The smallest model, ResNet-18 (\textbf{RN-18}), is selected for this study as the most suitable ResNet candidate to be used in embedded systems and has been extensively tested in continual learning studies \cite{hayes2020remind} \cite{rebuffi2017icarl, castro2018end, douillard2020podnet, wu2019large}.

\subsection{Online Continual Learning Models}

We assess how well the proposed model performs when compared to seven other online continuous learning techniques for training the classifier $F(\cdot)$ using the image feature vectors extracted using $G(\cdot)$. The techniques were selected because of low memory and computation requirements and they can learn incrementally, continuously, and with a single pass.

\textbf{Fine-Tune} incrementally adjusts a CNN's fully-connected layer. A stochastic gradient descent optimization strategy is used and progress is measured using cross-entropy loss of the CNN's predictions.

\textbf{Nearest Class Mean} (NCM) keeps a cumulative average vector for every unique class it encounters during training. Each class mean vector is considered a prototype representing a single class. During inference, NCM compares the input feature vector to the class mean vectors using a similarity metric such as Euclidean distance. The input is assigned to the class with the most similar feature mean vector. 

\textbf{Streaming One-vs-Rest} (SOvR) maintains a series of binary classifiers, one for each unique class it encounters during training. As each new feature vector continuously arrives in a streaming scenario, the classifier for the relevant class of the current input is updated incrementally. During inference, each classifier outputs a confidence score on whether the inferenced feature vector belongs to that class. The final predicted class is selected from the classifier with the best confidence score.

\textbf{Streaming Linear Discriminant Analysis} (SLDA) \cite{hayes2020lifelong} is an extension of Linear Discriminant Analysis designed to support learning from streaming data. The data distribution is modeled using class mean vectors and covariance matrices. A discriminant function is used to find a linear projection of the input data that maximizes the separation between classes. Both data distribution and discriminant function are updated incrementally as new feature vectors arrive from the data stream. During inference, SLDA uses the discriminant function to compute the probabilities of the inferenced vector belonging to each of the known classes. The final predicted class is selected from the class with the best probability score.

\textbf{Streaming Gaussian Naive Bayes} is an extension of the Gaussian Naive Bayes algorithm designed to support learning from streaming data. The model makes use of class-conditional probability distributions to determine if a feature vector belongs to a specific class. The distribution of each feature is represented by a mean and variance. The probability distributions are updated incrementally by observing the feature values of the incoming feature vectors from the data stream. During inference, Bayes' theorem is applied to obtain the posterior probability of each class based on the observed input feature. The predicted class is selected with the highest probability score.

\textbf{Online Perceptron} keeps a class vector for every unique class it encounters during training. When a feature vector is received, prediction is performed by taking the dot product of the input and the stored class vectors. The final predicted class is selected from the class with the best score. During training, no action is performed if the prediction matches the actual class. However if the prediction is a mismatch, the vector of the actual class is adjusted towards the input while the vector of the mismatched class is adjusted away from the input. This process is repeated continuously as the model receives additional feature vectors from the data stream.

\textbf{Replay} is a technique to reduce catastrophic forgetting by storing some of the previous training feature vectors in a memory replay buffer. During training, the model samples from incoming feature vectors equally alongside the stored feature vectors. Training examples can be selected from the buffer at random or by using specific strategies to mitigate issues such as imbalanced class representation. By incorporating past knowledge, the replay model balances the learning process to reduce catastrophic forgetting while giving equal attention to new knowledge. As training progresses, the memory buffer can be updated by replacing randomly-selected feature vectors with the current input, or by using specific strategies to retain important feature vectors. Replay can be memory intensive depending on how much storage is allocated for the memory buffer.

\textbf{Explainable Lifelong Learning} (ExLL) is the proposed model of this study. Three variations of the model were tested. \textbf{ExLL-P} uses Prototype-based Inference as per Equation \ref{eq_prinf} where label predictions are based on the closest prototype mean. \textbf{ExLL-M} uses MegaCloud-based Inference as per Equation \ref{eq_mcinf} where label predictions are based on the closest class mean. Lastly, \textbf{ExLL-F} uses pairwise fusion to combine the label predictions from ExLL-M and ExLL-P, as per Equation \ref{fusion_probability}.

\subsection{Datasets}

Online continual learners are evaluated using the following image classification datasets.

\textbf{OpenLORIS} \cite{she2020openloris} consists of videos of 40 different household items recorded from varying angles and distance, and 121 object instances across all items. Each object instance is recorded under one of the following environmental conditions: the object is surrounded by clutter; the object is illuminated by several light sources; the object is partially occluded; and the object is nearer to or further away from the camera. A total of 9 sessions are recorded for each condition and object instance. This dataset is suitable for testing a model's ability to learn and recognize objects from dynamic and sequential image streams as well as to apply its acquired knowledge to recognize known objects under different environments. 

\textbf{Places-365} \cite{zhou2017places} consists of 1.8 million images of locations divided into 365 categories. The dataset is segmented into a training and validation set. This dataset is suitable for evaluating a model's ability to learn from a large number of classes and diverse images per class. 

\textbf{Places-Long-Tail} (Places-LT) is a subset of Places-365 with a skewed distribution of images across all classes, designed to evaluate a model's ability to generalize from highly-imbalanced data distributions. Each of the 365 classes may consists of anywhere between 5 to thousands of images, while the validation set is identical to the validation set used by Places-365. 

\textbf{Few-Shot Incremental Object Learning} (F-SIOL-310) \cite{ayub2021fsiol} consists of static images of 22 household items. There are multiple instances of each item, totaling 310 object instances and 620 static images. This dataset uses two learning scenarios. The 5-shot learning scenario trains a model using only five images per class, selected at random, and tests with all other images. Likewise for the 10-shot learning scenario, only ten randomly-chosen images are used for training the model while all others are reserved as the testing set. Typically this dataset is used with multiple permutations of class orders and training images. This dataset is suitable for evaluating a model's ability to learn from few training samples.

\subsection{Experiment Protocol}

One of the factors impacting the performance of an online continual learner is the order in which training data is presented. This study presents several different orderings of each dataset and we observe the effects on the learners. 

Two variants of instance data orderings are used for the OpenLORIS dataset \cite{hayes2019memory}. \textbf{Instance ordering} shuffles object instances before presenting all training videos to the learner for training. On the other hand, \textbf{low-shot instance ordering} presents only one training video from each object instance and category to the learner during training. Having learned from the object instances, the learners are then tested on all testing videos of known objects. The low-shot ordering method evaluates how well the learner generalizes from a limited labeled dataset to identify known objects under various environmental conditions. 

For the two Places datasets, two data ordering methods are used. \textbf{Independent and identical distribution} (IID) shuffles the order of the images. \textbf{Class-IID} on the other hand organizes all the images by class; the order of the images is shuffled within each class, as well as the order of the classes. Class-IID is designed to test the learner's ability to handle catastrophic forgetting, and is commonly used as a continual learning metric \cite{rebuffi2017icarl} \cite{castro2018end} \cite{wu2019large} \cite{hou2019learning}. Some learners perform poorly with Class-IID ordering if they do not have catastrophic forgetting mitigation, but are still able to perform well using IID ordering.

Lastly, F-SIOL-310 is run using Class-IID ordering for each low-shot learning scenario. The experiment is run using three permutations of class orders and the averaged results are reported over all permutations.

\subsection{Performance Metric}

Online continual learners are evaluated on three axes: classification accuracy, number of parameters, and experiment runtime. The performance of an online learner $\mathcal{M}$ is computed as a modified NetScore metric \cite{wong2019netscore} combining all three metrics into one score as follows:
\begin{equation} \label{eqn_netscore}
    \Omega (\mathcal{M}) = s \log (\frac{a (\mathcal{M})^{\alpha}}{p (\mathcal{M})^{\beta} c (\mathcal{M})^{\gamma}})
\end{equation}
where $a(\mathcal{M})$ is the learner's testing accuracy, $p(\mathcal{M})$ is the learner's size, $c(\mathcal{M})$ is the time taken to complete the experiment from start to finish, and $\alpha, \beta, \gamma$ are user-defined constants for controlling the contributions of accuracy, number of parameters, and the experiment runtime towards computing the NetScore $\Omega$.

The NetScore parameters follow the original parameter settings as suggested by Hayes et al. \cite{hayes2022online}. $s=20$ and $\alpha=2$ to prioritize classification accuracy, and $\beta = \gamma = 0.25$ to moderate the large values of $p(\mathcal{M})$ and $c(\mathcal{M})$ \cite{wong2019netscore}. Higher NetScores indicate better performance. 

\section{Results}
% Results for each Dataset
% Present Table 1 (NetScore)
% Present learning curve
% Present spiderweb graph
% Explainable rules 
% Discussion

For OpenLORIS and Places-LT, the results are reported from the average performance across three permutations for each data ordering technique. On the other hand, Places-365 is run only once for each ordering due to the long time needed to complete the experiment. Meanwhile, classifiers such as SOvR and NCM are relatively unaffected by data ordering permutations due to the usage of running class mean vectors. As for the Replay method, two buffer sizes were compared: one storing 20 training samples per class (20pc) and the other storing 2 training samples per class (2pc).

\subsection{Results on OpenLORIS}

\begin{figure}
    \centering
    \includegraphics[width=0.99\textwidth]{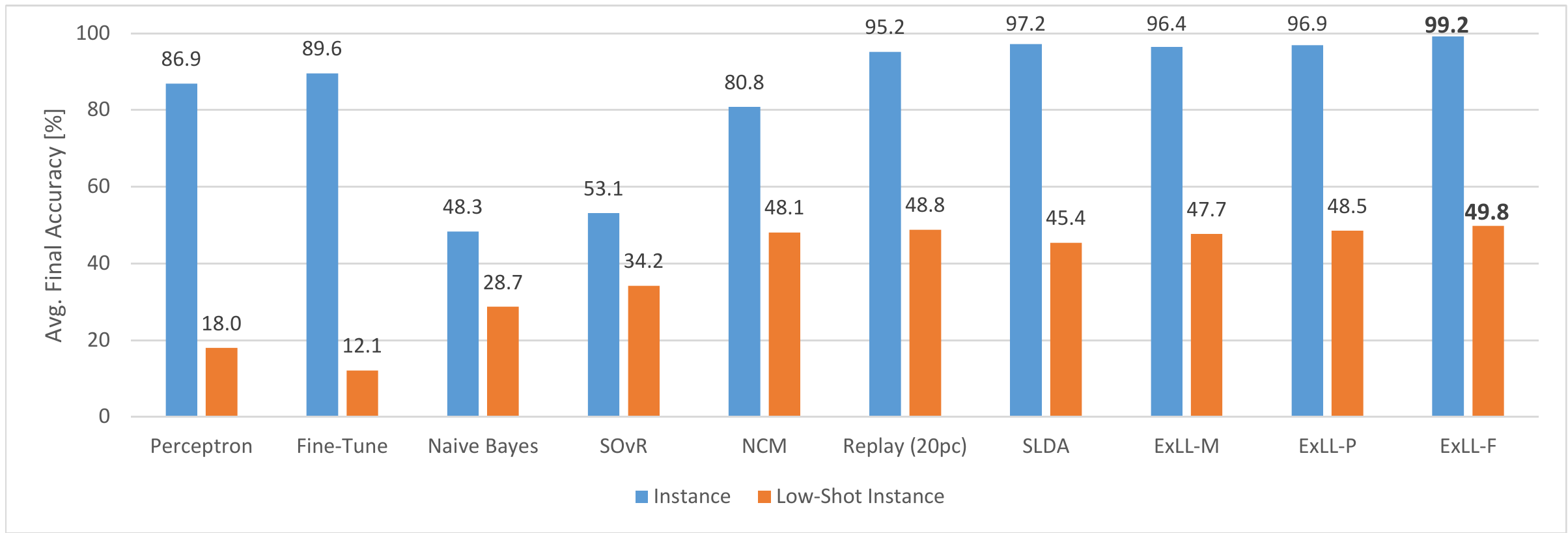}
    \caption{Accuracy results averaged across CNN architectures comparing online continual learners' performance on OpenLORIS with \textbf{instance ordering} and \textbf{low-shot instance ordering}.}
    \label{fig_barresults_openloris_mean}
\end{figure}

Online continual learners are evaluated on OpenLORIS using two data ordering methods. \textbf{Instance ordering} trains learners on all object instances while \textbf{low-shot instance ordering} trains learners on one object instance from each object class. Performance is evaluated by combining the top-1 accuracy scores of the top-ranked choice for each learner. The scores are then averaged across all CNN architectures to compare how various orderings affect learner performance, as shown in Figure \ref{fig_barresults_openloris_mean}. 

All models displayed lower accuracy when using low-shot instance ordering. Perceptron and Fine-Tune in particular showed a much bigger drop in accuracy for the low-shot instance ordering, relative to other continual learning models. The models generalized poorly when tested against images from domains not encountered during training. In comparison, Naive Bayes, SOvR, and NCM were less accurate than Perceptron and Fine-Tune for the full instance ordering, but outperformed them for the low-shot condition. The ExLL models showed the best balance between the two ordering methods, while ExLL-F outperformed all other models for both orderings. 

\begin{table}[]
    \centering
    \caption{NetScores on OpenLORIS with the \textbf{low-shot instance ordering}. \textbf{Higher} values are better. Results are highlighted as follows for the \textbf{\underline{first}}, \textbf{second}, and \underline{third} best results.}
    \label{tab_results_openloris_netscore1}
    \begin{tabular}{c|c c c c c|c}
        \hline 
        Method & MNet-S & MNet-L & ENet-B0 & ENet-B1 & RN-18 & Mean \\ \hline
        Perceptron      & -115.9 & -106.5 & -91.0 & -96.5 & -147.8 & -111.5 \\
        Fine-Tune       & -149.0 & -142.9 & -96.3 & -103.7 & -187.8 & -135.9 \\
        Naive Bayes     & -83.7 & -77.3 & -75.8 & -84.2 & -204.5 & -105.1 \\
        SOvR & -80.0    & -83.9 & -74.0 & \underline{-78.7} & -111.4 & -85.6 \\
        NCM & \textbf{\underline{-55.5}} & \textbf{\underline{-64.7}} & \textbf{\underline{-65.3}} & \textbf{\underline{-72.5}} & \textbf{\underline{-78.7}} & \textbf{\underline{-67.3}} \\
        Replay (20pc)   & \underline{-58.9} & \textbf{-66.5} & \textbf{-66.6} & \textbf{-72.8} & \textbf{-80.5} & \textbf{-69.1} \\
        SLDA & \textbf{-58.3} & \underline{-69.1} & \underline{-72.3} & -79.3 & \underline{-80.8} & \underline{-72.0} \\ \hline
        \textbf{ExLL}   & -75.3 & -91.4 & -99.5 & -106.8 & -108.1 & -96.2 \\
        \hline
    \end{tabular}
\end{table}

\subsubsection{NetScore Performance}

In Table \ref{tab_results_openloris_netscore1}, NetScores were used for evaluating continual learning methods in terms of performance as well as memory and computational requirements. The NetScore values were obtained by evaluating all methods on the same hardware for consistency. Higher Netscore values are better. 

The top three performing online continual learners are NCM, Replay 20pc, and SLDA. The NCM algorithm is the most efficient in terms of memory and computation requirements since it only stores and updates the class mean vectors. Replay 20pc needed additional computation and memory for replaying stored samples, while SLDA needed additional computation and memory for the covariance matrix. Meanwhile, ExLL was placed fifth among the eight algorithms. While ExLL is nominally similar to SLDA, ExLL required significantly more memory to store prototype mean vectors in addition to class mean vectors. In addition, ExLL stores training records to be able to recall the information for explaining inferences. 

\begin{table}[]
    \centering
    \caption{Accuracy results on OpenLORIS with the \textbf{instance ordering}. Results are highlighted as follows for the \textbf{\underline{first}}, \textbf{second}, and \underline{third} best results.}
    \begin{tabular}{c|c c c c c|c}
        \hline
        Method & MNet-S & MNet-L & ENet-B0 & ENet-B1 & RN-18 & Mean \\ \hline
        Perceptron & 0.793 & 0.880 & 0.935 & 0.942 & 0.796 & 0.869 \\
        Fine-Tune & 0.835 & 0.915 & 0.958 & 0.963 & 0.809 & 0.896 \\
        Naive Bayes & 0.311 & 0.526 & 0.780 & 0.787 & 0.015 & 0.483 \\
        SOvR & 0.374 & 0.477 & 0.739 & 0.723 & 0.346 & 0.531 \\
        NCM & 0.729 & 0.789 & 0.859 & 0.867 & 0.797 & 0.808 \\
        % Replay (2pc) &  &  &  &  &  &  \\
        Replay (20pc) & 0.921 & 0.956 & 0.977 & 0.978 & 0.929 & 0.952 \\
        SLDA & \textbf{0.956} & \textbf{0.982} & \textbf{0.988} & \textbf{0.988} & \underline{0.950} & \textbf{0.972} \\ \hline
        \textbf{ExLL-M} & 0.944 & \underline{0.973} & \underline{0.982} & 0.982 & 0.940 & 0.964 \\ 
        \textbf{ExLL-P} & \underline{0.951} & 0.968 & \underline{0.982} & \underline{0.983} & \textbf{0.961} & \underline{0.969} \\
        \textbf{ExLL-F} & \textbf{\underline{0.987}} & \textbf{\underline{0.993}} & \textbf{\underline{0.996}} & \textbf{\underline{0.996}} & \textbf{\underline{0.988}} & \textbf{\underline{0.992}} \\ \hline
    \end{tabular}
    \label{tab_results_openloris_acc1}
\end{table}

\begin{table}[]
    \centering
    \caption{Accuracy results on OpenLORIS with the \textbf{low-shot instance ordering}. Results are highlighted as follows for the \textbf{\underline{first}}, \textbf{second}, and \underline{third} best results.}
    \begin{tabular}{c|c c c c c|c}
        \hline
        Method & MNet-S & MNet-L & ENet-B0 & ENet-B1 & RN-18 & Mean \\ \hline
        Perceptron & 0.098 & 0.167 & 0.272 & 0.283 & 0.082 & 0.180 \\
        Fine-Tune & 0.043 & 0.066 & 0.238 & 0.232 & 0.030 & 0.121 \\
        Naive Bayes & 0.232 & 0.366 & 0.421 & 0.399 & 0.021 & 0.287 \\
        SOvR & 0.259 & 0.323 & 0.449 & 0.459 & 0.224 & 0.342 \\
        NCM & 0.442 & 0.474 & \textbf{0.516} & \textbf{0.514} & \underline{0.463} & 0.481 \\
        Replay (20pc) & 0.453 & 0.480 & \textbf{\underline{0.529}} & \textbf{\underline{0.532}} & 0.446 & \textbf{0.488} \\
        SLDA & 0.445 & 0.454 & 0.472 & 0.460 & 0.442 & 0.454 \\ \hline
        \textbf{ExLL-M} & \underline{0.463} & \underline{0.493} & 0.504 & 0.487 & 0.440 & 0.477 \\
        \textbf{ExLL-P} & \textbf{0.470} & \textbf{0.501} & 0.500 & 0.482 & \textbf{0.475} & \underline{0.485} \\
        \textbf{ExLL-F} & \textbf{\underline{0.481}} & \textbf{\underline{0.511}} & \underline{0.511} & \underline{0.495} & \textbf{\underline{0.492}} & \textbf{\underline{0.498}} \\ \hline
    \end{tabular}
    \label{tab_results_openloris_acc2}
\end{table}

\subsubsection{Backbone CNN Comparisons}

Table \ref{tab_results_openloris_acc1} presents the best accuracy scores of online continual learning models across different CNN backbones when trained using instance ordering. The EfficientNet architectures showed the best results overall while ResNet-18 showed the worst results in eight of the ten datasets. 

Table \ref{tab_results_openloris_acc2} presents the performance of the models across different CNN architectures for low-shot instance ordering. Compared to the previous table, classification accuracy was significantly lower due to fewer training samples. The EfficientNet backbone CNNs again outperformed the other backbone CNNs.

\begin{table}[]
    \centering
    \caption{Accuracy results on Places-365 for two data ordering methods \textbf{iid} and \textbf{class-iid}. Results are highlighted as follows for the \textbf{\underline{first}}, \textbf{second}, and \underline{third} best results.}
    \begin{tabular}{c | c c c c c | c c c c c | c}
        \hline
        \multirow{2}{*}{Method} & \multicolumn{5}{c|}{IID} & \multirow{2}{*}{Mean} \\ 
         & MNet-S & MNet-L & ENet-B0 & ENet-B1 & RN-18 & \\ \hline
         Perceptron & 0.303 & 0.344 & 0.352 & 0.340 & 0.294 & 0.326 \\
         Fine-Tune & 0.214 & 0.252 & 0.293 & 0.280 & 0.217 & 0.251 \\
         Naive Bayes & 0.028 & 0.093 & 0.250 & 0.249 & 0.003 & 0.124 \\
         NCM & 0.285 & 0.332 & 0.361 & 0.356 & 0.322 & 0.331 \\
         Replay (20pc) & 0.289 & 0.323 & 0.354 & 0.348 & 0.261 & 0.315 \\
         SLDA & \underline{0.362} & \textbf{0.397} & \textbf{0.412} & \textbf{0.405} & \textbf{0.362} & \textbf{0.387} \\ \hline
         \textbf{ExLL-M} & \textbf{0.375} & 0.347 & \underline{0.392} & 0.336 & 0.312 & 0.352 \\
         \textbf{ExLL-P} & 0.354 & \underline{0.380} & 0.381 & \underline{0.370} & \underline{0.343} & \underline{0.365} \\
         \textbf{ExLL-F} & \textbf{\underline{0.444}} & \textbf{\underline{0.478}} & \textbf{\underline{0.488}} & \textbf{\underline{0.476}} & \textbf{\underline{0.440}} & \textbf{\underline{0.465}} \\ \hline
    \end{tabular}

    \begin{tabular}{c}
    \\
    \end{tabular}

    \begin{tabular}{c | c c c c c | c}
        \hline
        \multirow{2}{*}{Method} & \multicolumn{5}{c|}{Class-IID} & \multirow{2}{*}{Mean} \\ 
         & MNet-S & MNet-L & ENet-B0 & ENet-B1 & RN-18 & \\ \hline
         Perceptron & 0.004 & 0.003 & 0.012 & 0.013 & 0.005 & 0.007 \\
         Fine-Tune & 0.003 & 0.003 & 0.006 & 0.006 & 0.003 & 0.004 \\
         Naive Bayes & 0.028 & 0.093 & 0.250 & 0.249 & 0.003 & 0.124 \\
         NCM & 0.265 & 0.309 & 0.336 & 0.329 & 0.300 & 0.307 \\
         Replay (20pc) & 0.251 & 0.279 & 0.297 & 0.295 & 0.235 & 0.271 \\
         SLDA & \textbf{0.362} & \textbf{0.397} & \textbf{0.412} & \textbf{0.405} & \textbf{0.362} & \textbf{0.387} \\ \hline
         \textbf{ExLL-M} & 0.347 & 0.362 & \underline{0.381} & 0.367 & \underline{0.349} & 0.361 \\
         \textbf{ExLL-P} & \underline{0.352} & \underline{0.378} & \underline{0.381} & \underline{0.373} & 0.343 & \underline{0.365} \\
         \textbf{ExLL-F} & \textbf{\underline{0.444}} & \textbf{\underline{0.473}} & \textbf{\underline{0.486}} & \textbf{\underline{0.479}} & \textbf{\underline{0.439}} & \textbf{\underline{0.464}} \\ \hline
    \end{tabular}
    
    \label{tab_results_places_acc2}
\end{table}

\begin{table}[]
    \centering
    \caption{Accuracy results on Places-LT for two data ordering methods \textbf{iid} and \textbf{class-iid}. Accuracy scores are averaged over three runs with different data permutations. Results are highlighted as follows for the \textbf{\underline{first}}, \textbf{second}, and \underline{third} best results.}
    \begin{tabular}{c | c c c c c | c }
        \hline
        \multirow{2}{*}{Method} & \multicolumn{5}{c|}{IID} & \multirow{2}{*}{Mean} \\ 
         & MNet-S & MNet-L & ENet-B0 & ENet-B1 & RN-18 & \\ \hline
         Perceptron & 0.152 & 0.185 & 0.213 & 0.206 & 0.149 & 0.181 \\
         Fine-Tune & 0.136 & 0.163 & 0.197 & 0.191 & 0.141 & 0.165\\
         Naive Bayes & 0.015 & 0.050 & 0.199 & 0.213 & 0.100 & 0.115\\
         SOvR & 0.089 & 0.149 & 0.262 & 0.245 & 0.146 & 0.178\\
         NCM & 0.265 & 0.309 & 0.336 & \underline{0.329} & \underline{0.300} & 0.306\\
         Replay (20pc) & 0.239 & 0.267 & 0.290 & 0.282 & 0.223 & 0.260 \\
         SLDA & \underline{0.290} & \underline{0.318} & \underline{0.338} & 0.328 & \underline{0.300} & \underline{0.315} \\ \hline
         \textbf{ExLL-M} & \textbf{\underline{0.356}} & \textbf{\underline{0.392}} & \textbf{\underline{0.407}} & \textbf{\underline{0.400}} & \textbf{\underline{0.360}} & \textbf{\underline{0.383}} \\
         \textbf{ExLL-P} & 0.265 & 0.297 & 0.315 & 0.305 & 0.277 & 0.292 \\
         \textbf{ExLL-F} & \textbf{0.324} & \textbf{0.349} & \textbf{0.362} & \textbf{0.351} & \textbf{0.331} & \textbf{0.343} \\ \hline
    \end{tabular}

    \begin{tabular}{c}
    \\
    \end{tabular}

    \begin{tabular}{c | c c c c c | c}
        \hline
        \multirow{2}{*}{Method} & \multicolumn{5}{c|}{Class-IID} & \multirow{2}{*}{Mean} \\ 
         & MNet-S & MNet-L & ENet-B0 & ENet-B1 & RN-18 & \\ \hline
         Perceptron & 0.017 & 0.028 & 0.071 & 0.073 & 0.015 & 0.041 \\
         Fine-Tune & 0.015 & 0.021 & 0.071 & 0.075 & 0.004 & 0.037\\
         Naive Bayes & 0.015 & 0.050 & 0.199 & 0.213 & 0.001 & 0.096\\
         SOvR & 0.089 & 0.149 & 0.262 & 0.245 & 0.146 & 0.178\\
         NCM & 0.265 & 0.309 & 0.336 & \underline{0.329} & \underline{0.300} & 0.308 \\
         Replay (20pc) & 0.241 & 0.268 & 0.306 & 0.295 & 0.193 & 0.261 \\
         SLDA & \underline{0.290} & \underline{0.319} & \textbf{0.338} & 0.328 & \underline{0.300} & \underline{0.315} \\ \hline
         \textbf{ExLL-M} & \textbf{0.311} & \textbf{0.331} & \textbf{0.338} & \textbf{0.331} & \textbf{\underline{0.345}} & \textbf{0.331} \\
         \textbf{ExLL-P} & 0.265 & 0.300 & 0.314 & 0.303 & 0.273 & 0.291 \\
         \textbf{ExLL-F} & \textbf{\underline{0.325}} & \textbf{\underline{0.351}} & \textbf{\underline{0.359}} & \textbf{\underline{0.350}} & \textbf{0.330} & \textbf{\underline{0.343}}\\ \hline
    \end{tabular}
    
    \label{tab_results_places_acc3}
\end{table}

\subsection{Results on Places-365 and Places-LT}

In this section, online continual learners were compared in terms of performance, regardless of which CNN architecture was used. Tables \ref{tab_results_places_acc2} and \ref{tab_results_places_acc3} show the average top-1 accuracy for Places-365 and Places-LT, respectively, for all online continual learners across all CNNs. In almost every case, SLDA outperformed ExLL-M and ExLL-P. 

However, when pairwise fusion was used for combining the results from the two ExLL methods, ExLL-F was able to outperform SLDA, ExLL-M, and ExLL-P by a significant margin. This suggests that the local and global inferences in ExLL-F contain complementary information and were able to address each other's weaknesses when pairwise fusion is used.

Perceptron and Fine-Tune show a significant drop in accuracy in Class-IID compared to IID, due to catastrophic forgetting. When training using Class-IID, known classes are not revisited and are thus negatively impacted when new classes are introduced. Other online continual learning models, including ExLL, are not as affected by catastrophic forgetting. 

Both Places datasets use the same set of images for testing but with different training sets. While Places-365 tests generalization for 365 location-based classes with 1.8 million images, Places-LT tests how well models perform with severe imbalance, with classes consisting of anywhere between 5 to 4,980 training images. Therefore, comparing the performance of the models for the two datasets is one way to observe their robustness against dataset imbalance. Of the three ExLL methods, MegaCloud-based inference was the least affected by dataset imbalance while prototype-based inference and pairwise fusion showed a 7.4\% and 12.1\% loss in performance respectively when trained with Places-LT. ExLL-F in particular showed worse performance compared to either ExLL-M and ExLL-P, demonstrating a significant vulnerability to imbalance. 

A visualization of the topology of prototypes is provided as a supplementary material (Figure \ref{fig_visualize_places}).

% Figure \ref{fig_visualize_places} visualizes the topology of the prototypes encoded using ExLL-F. In Sub-Figure \ref{fig_visualize_places}(a), the high-dimensional centroids are transformed into 2-dimensional scatter plots using t-distributed stochastic neighbor embedding (t-SNE) \cite{van2008visualizing}. Each circle denotes one prototype and the radius of the circle indicates the support or size of the prototype's data cloud. There are multiple instances where neighbouring prototypes displayed overlapping areas of influences. The presence of overlap suggests that some prototypes are highly similar to each other even after self-organization. It is possible to reduce redundancies by merging the overlapping prototypes, for example, by using hierarchical clustering \cite{wang2012hierarchical} or divide-and-merge \cite{rehman2022divide}. In Sub-Figure \ref{fig_visualize_places}(b), the same 2-dimensional prototype topology is represented using a Voronoi tesselation graph. The merged prototypes, for instance, may be visualized by removing the edges between neighbouring partitions with the same colour (i.e., same class labels). 

\begin{table}[]
    \centering
    \caption{Accuracy results on F-SIOL-310 using \textbf{class-iid data ordering} for \textbf{5-shot} learning and \textbf{10-shot} learning scenarios. Accuracy scores are averaged over three runs with different data permutations. Results are highlighted as follows for the \textbf{\underline{first}}, \textbf{second}, and \underline{third} best results.}
    
    \begin{tabular}{c | c c c c c c}
        \hline
        \multirow{2}{*}{Method} & \multicolumn{6}{c}{5-Shot} \\ 
         & MNet-S & MNet-L & ENet-B0 & ENet-B1 & RN-18 & Mean \\ \hline
         Perceptron & 0.181 & 0.177 & 0.406 & 0.454 & 0.049 & 0.253 \\
         Fine-Tune & 0.183 & 0.205 & 0.416 & 0.460 & 0.090 & 0.270 \\
         Naive Bayes & 0.344 & 0.554 & 0.816 & 0.828 & 0.035 & 0.515 \\
         SOvR & 0.592 & 0.666 & 0.679 & 0.693 & 0.428 & 0.611 \\
         CBCL & \underline{0.853} & \underline{0.878} & \underline{0.886} & 0.838 & 0.848 & 0.860 \\
         NCM & \underline{0.853} & 0.871 & \underline{0.886} & \textbf{0.885} & \textbf{\underline{0.885}} & \underline{0.876} \\
         Replay (20pc) & 0.541 & 0.632 & 0.594 & 0.612 & 0.624 & 0.600 \\
         SLDA & \textbf{0.880} & \textbf{0.899} & \textbf{\underline{0.912}} & \textbf{\underline{0.903}} & \underline{0.854} & \textbf{\underline{0.889}} \\ \hline
         \textbf{ExLL-M} & 0.842 & 0.873 & 0.863 & 0.847 & 0.851 & 0.855 \\
         \textbf{ExLL-P} & 0.827 & 0.832 & 0.799 & 0.755 & 0.803 & 0.803 \\
         \textbf{ExLL-F} & \textbf{\underline{0.889}} & \textbf{\underline{0.905}} & \textbf{0.887} & \underline{0.854} & \textbf{\underline{0.885}} & \textbf{0.884} \\ \hline
    \end{tabular}

    \begin{tabular}{c}
    \\
    \end{tabular}

    \begin{tabular}{c | c c c c c c }
        \hline
        \multirow{2}{*}{Method} & \multicolumn{6}{|c}{10-Shot} \\ 
         & MNet-S & MNet-L & ENet-B0 & ENet-B1 & RN-18 & Mean \\ \hline
         Perceptron & 0.158 & 0.223 & 0.354 & 0.458 & 0.051 & 0.248 \\
         Fine-Tune & 0.127 & 0.199 & 0.389 & 0.453 & 0.090 & 0.251 \\
         Naive Bayes & 0.320 & 0.537 & 0.806 & 0.854 & 0.015 & 0.506 \\
         SOvR & 0.561 & 0.702 & 0.650 & 0.752 & 0.504 & 0.633 \\
         CBCL & 0.883 & 0.906 & 0.888 & 0.892 & 0.869 & 0.887 \\
         NCM & 0.883 & 0.906 & 0.893 & 0.913 & 0.896 & 0.898 \\
         Replay (20pc) & 0.625 & 0.694 & 0.714 & 0.722 & 0.731 & 0.697 \\
         SLDA & 0.924 & \textbf{0.948} & \textbf{0.938} & \textbf{0.936} & 0.910 & \underline{0.931} \\ \hline
         \textbf{ExLL-M} & \underline{0.926} & \underline{0.942} & \textbf{0.938} & \underline{0.928} & \textbf{0.948} & \textbf{0.936} \\
         \textbf{ExLL-P} & \textbf{0.927} & 0.934 & 0.897 & 0.879 & \underline{0.930} & 0.913 \\
         \textbf{ExLL-F} & \textbf{\underline{0.961}} & \textbf{\underline{0.966}} & \textbf{\underline{0.952}} & \textbf{\underline{0.943}} & \textbf{\underline{0.968}} & \textbf{\underline{0.958}} \\ \hline
    \end{tabular}
    \label{tab_results_fsiol_acc1}
\end{table}

\subsection{Results on F-SIOL-310}

F-SIOL-310 was selected to observe how the online continual learning methods perform in low-shot continuous learning applications. Table \ref{tab_results_fsiol_acc1} presents the performance for all continual learning methods, backbone CNNs, and learning scenarios. For the 5-shot scenario, ExLL-F is slightly outperformed by SLDA (0.884 vs. 0.889 respectively). On the other hand, for the 10-shot scenario, ExLL-F significantly outperformed the next-best methods, ExLL-M and SLDA (0.958 vs. 0.936 and 0.931 respectively). 

A visualization of the topology of prototypes is provided as a supplementary material (Figure \ref{fig_visualize_fsiol}).

\begin{figure*}[]
    \centering
    \includegraphics[width=0.24\linewidth]{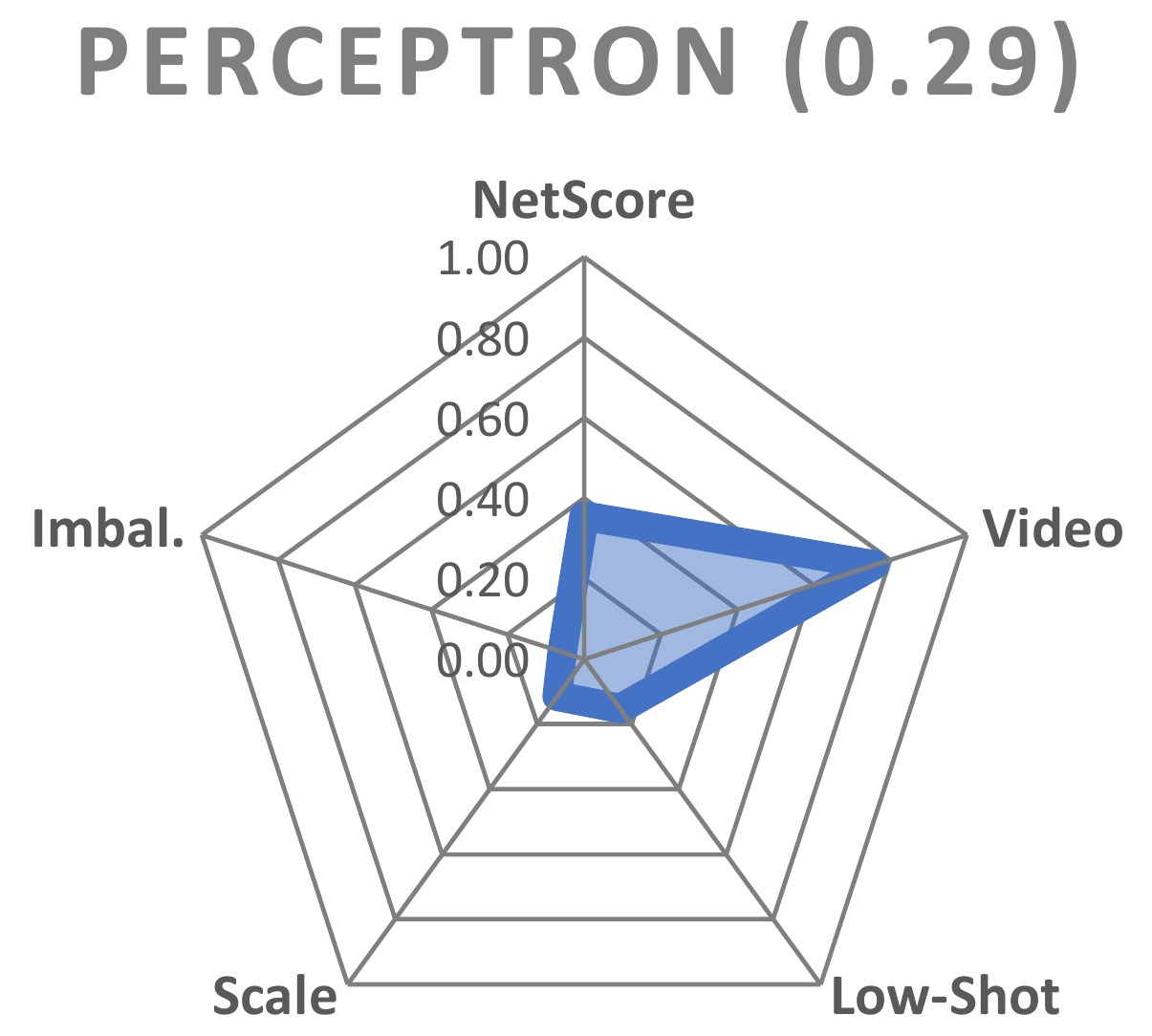}
    \includegraphics[width=0.24\linewidth]{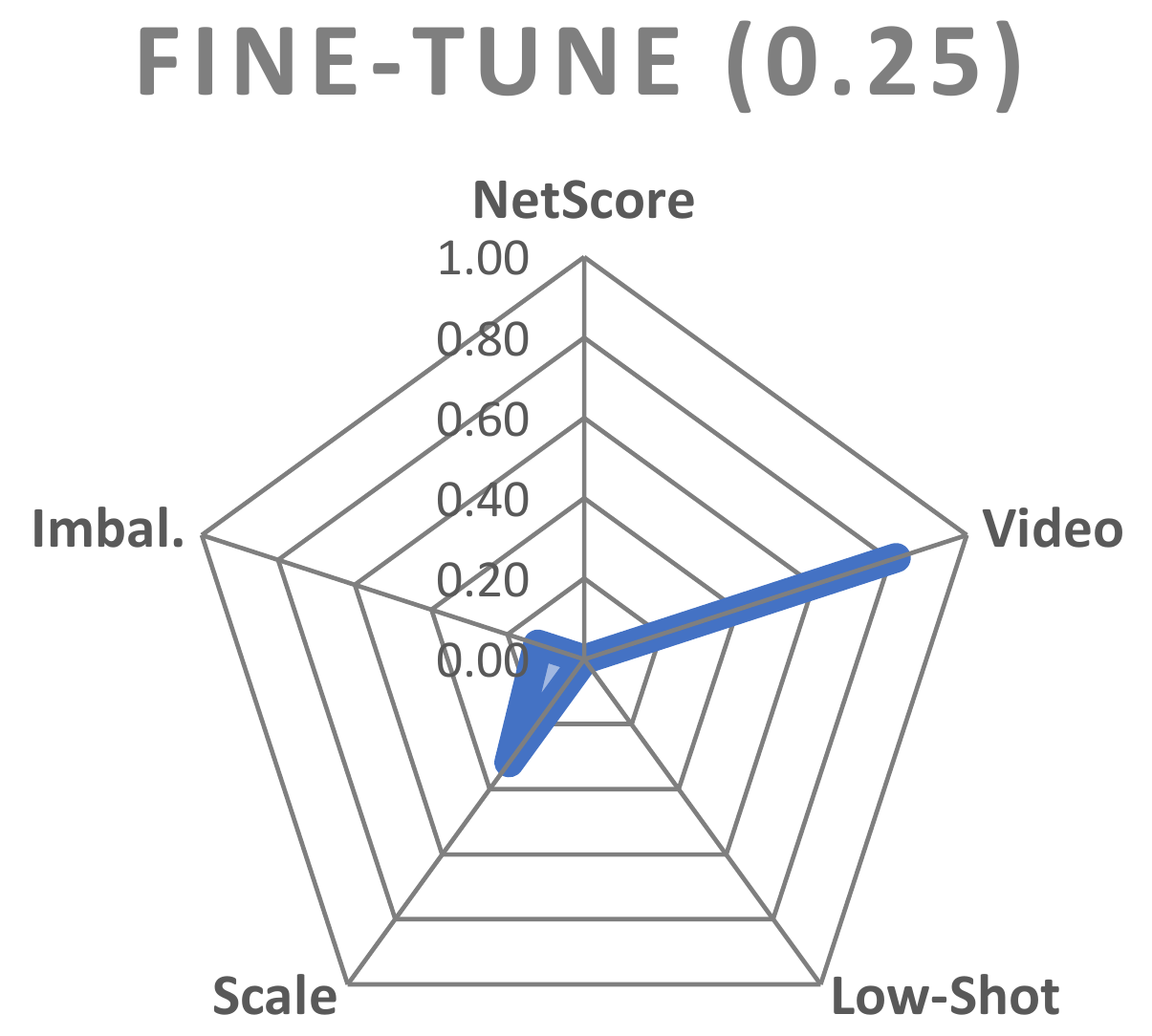}
    \includegraphics[width=0.24\linewidth]{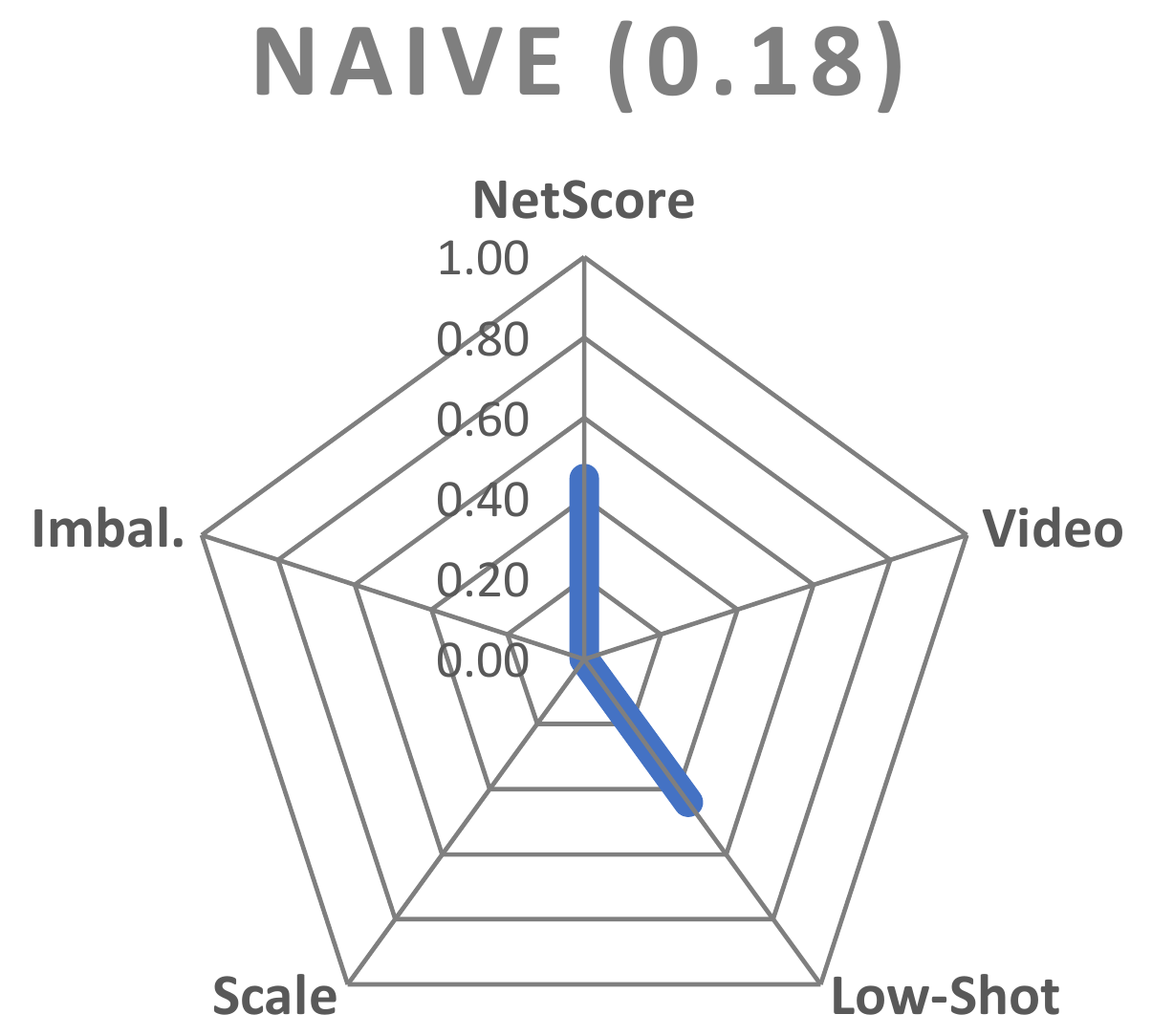}
    \includegraphics[width=0.24\linewidth]{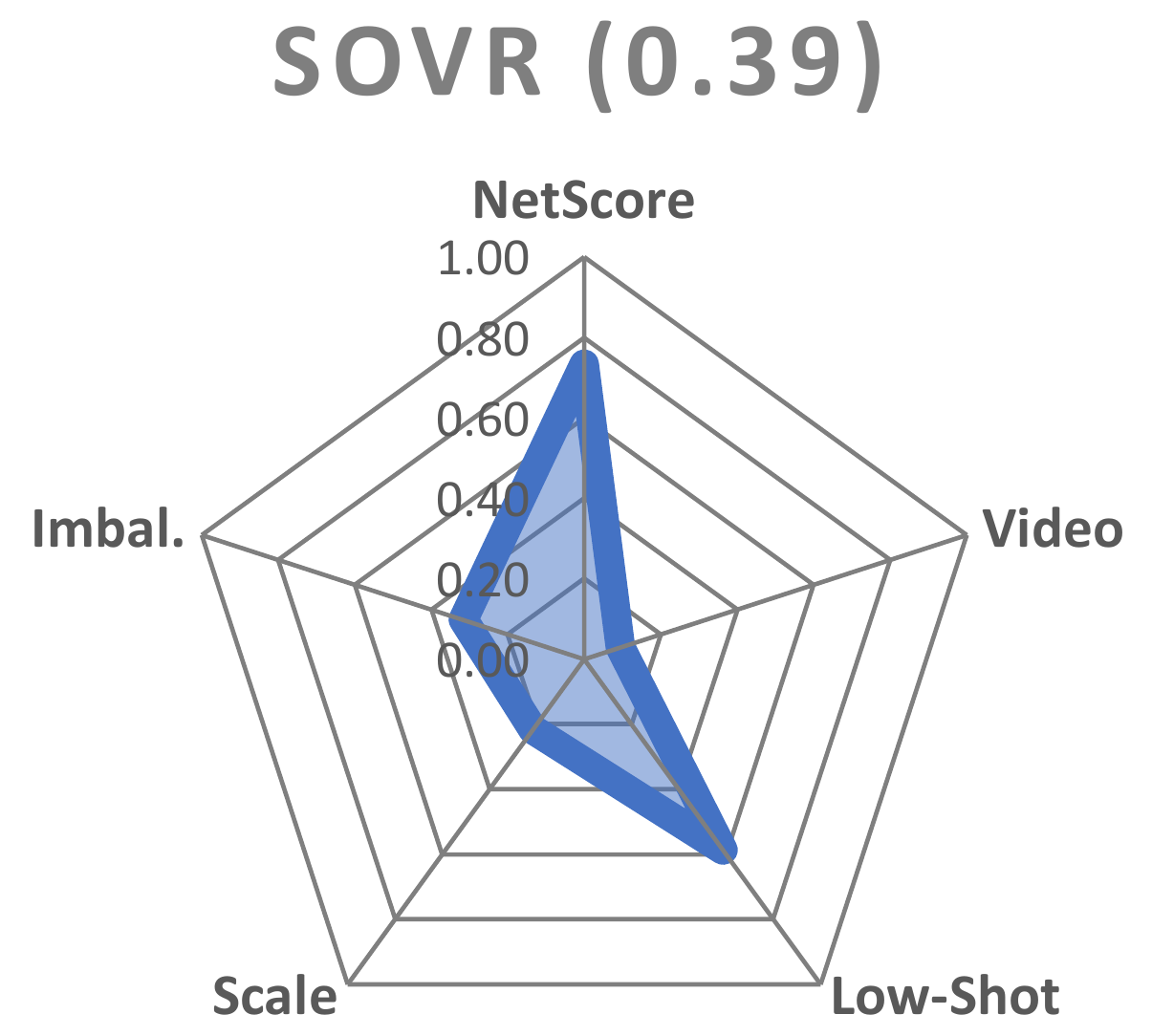} 
    
    \hfill
    
    \includegraphics[width=0.24\linewidth]{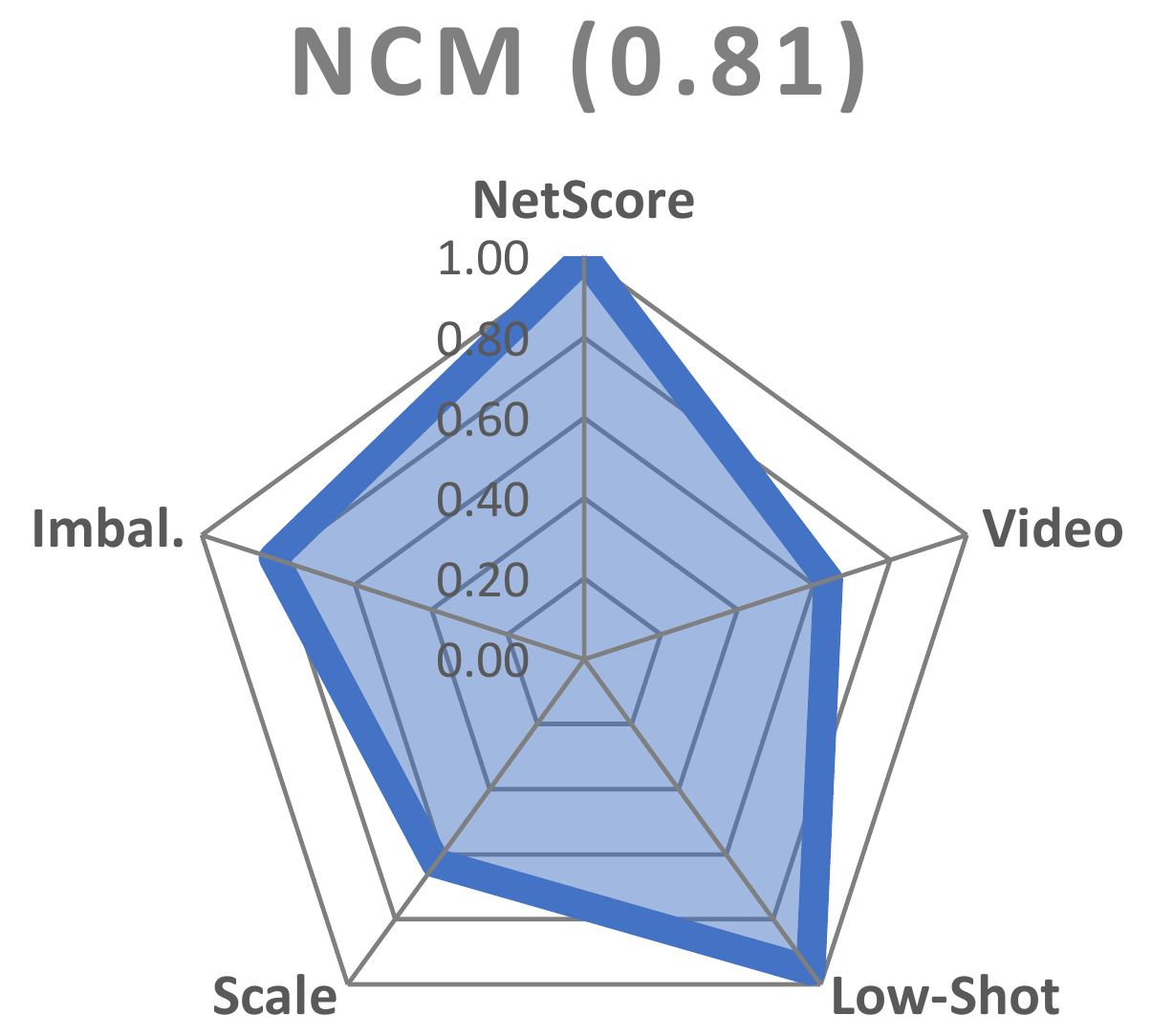}
    \includegraphics[width=0.24\linewidth]{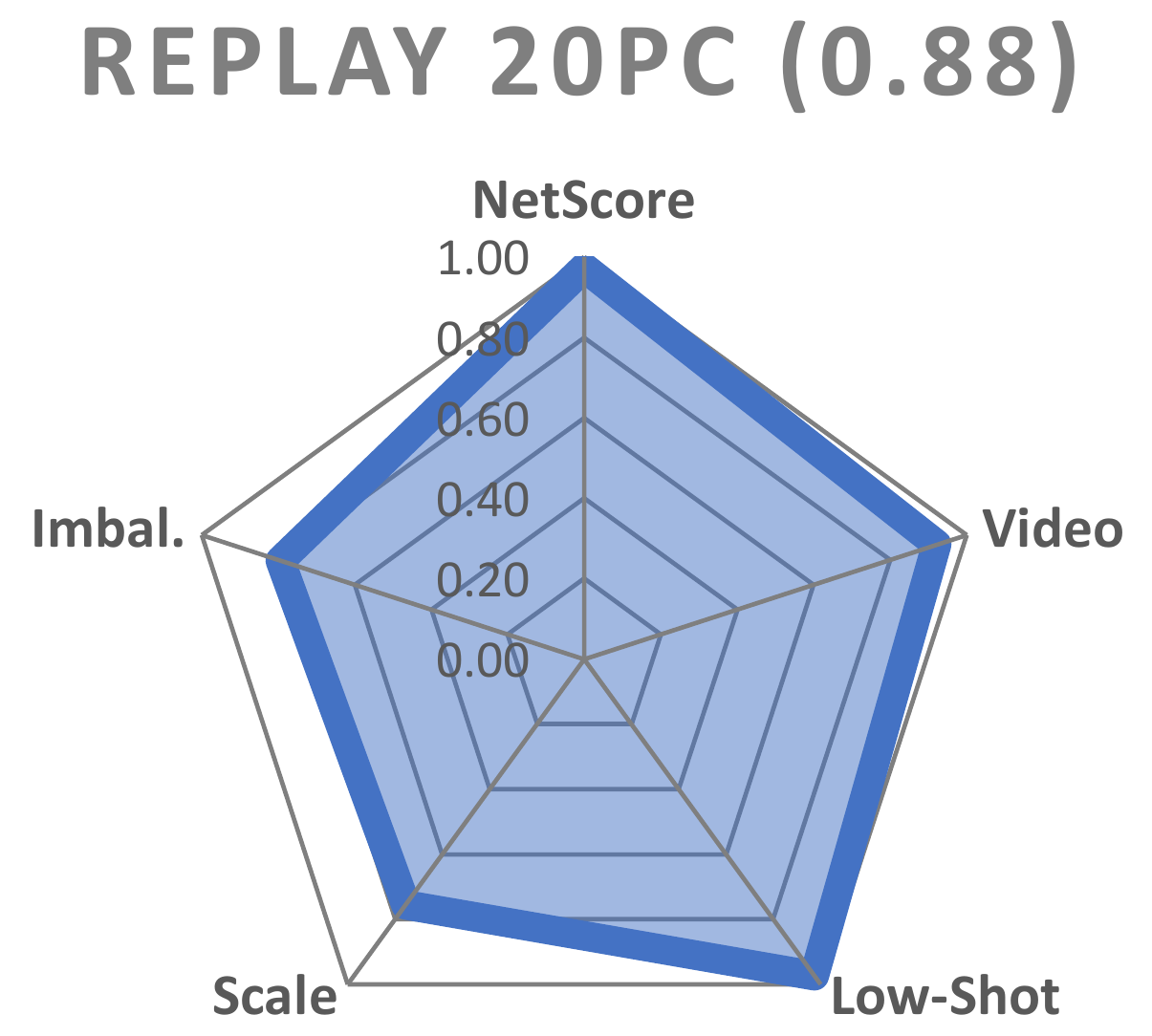}
    \includegraphics[width=0.24\linewidth]{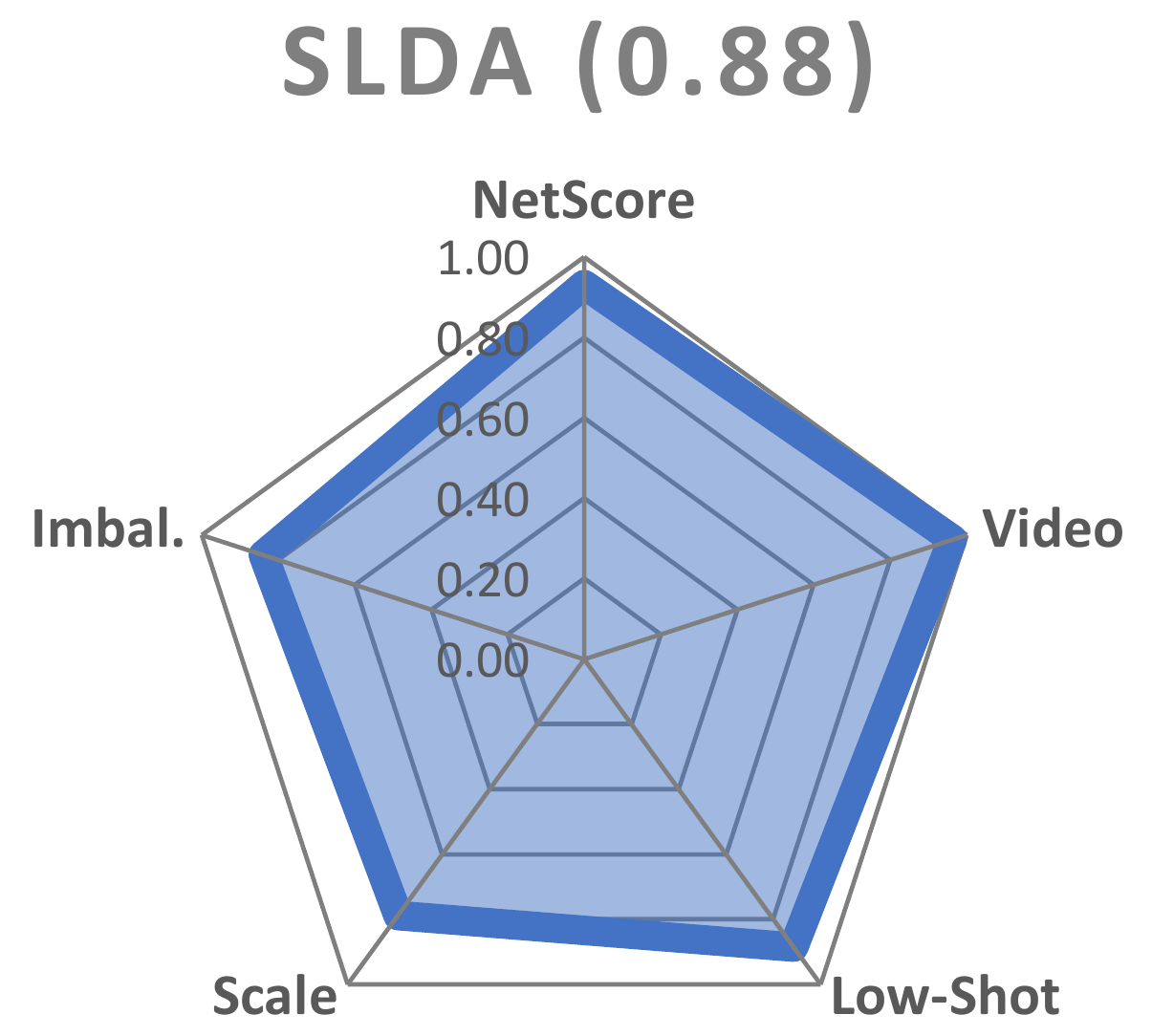} 

    \hfill
    
    \includegraphics[width=0.24\linewidth]{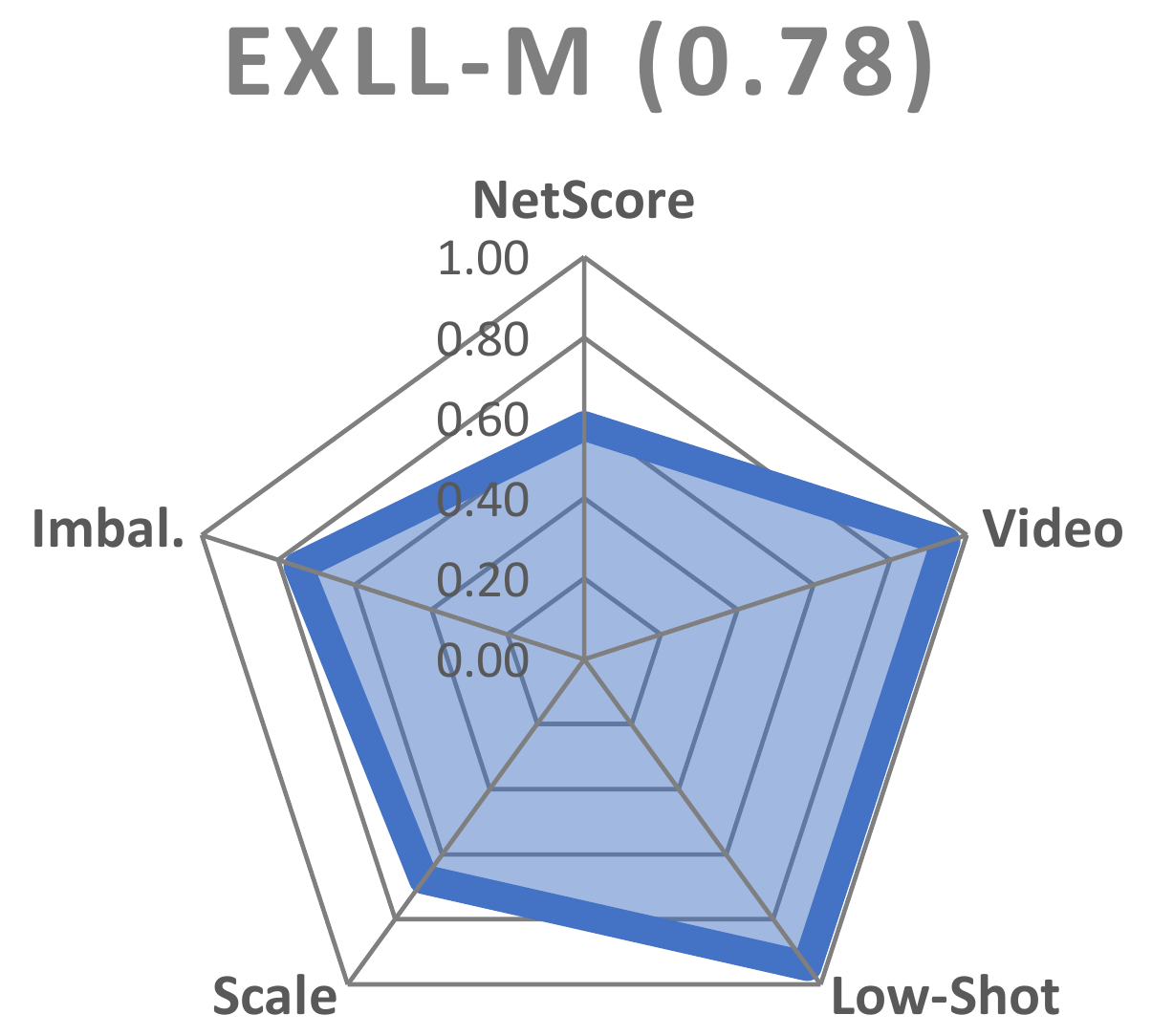}
    \includegraphics[width=0.24\linewidth]{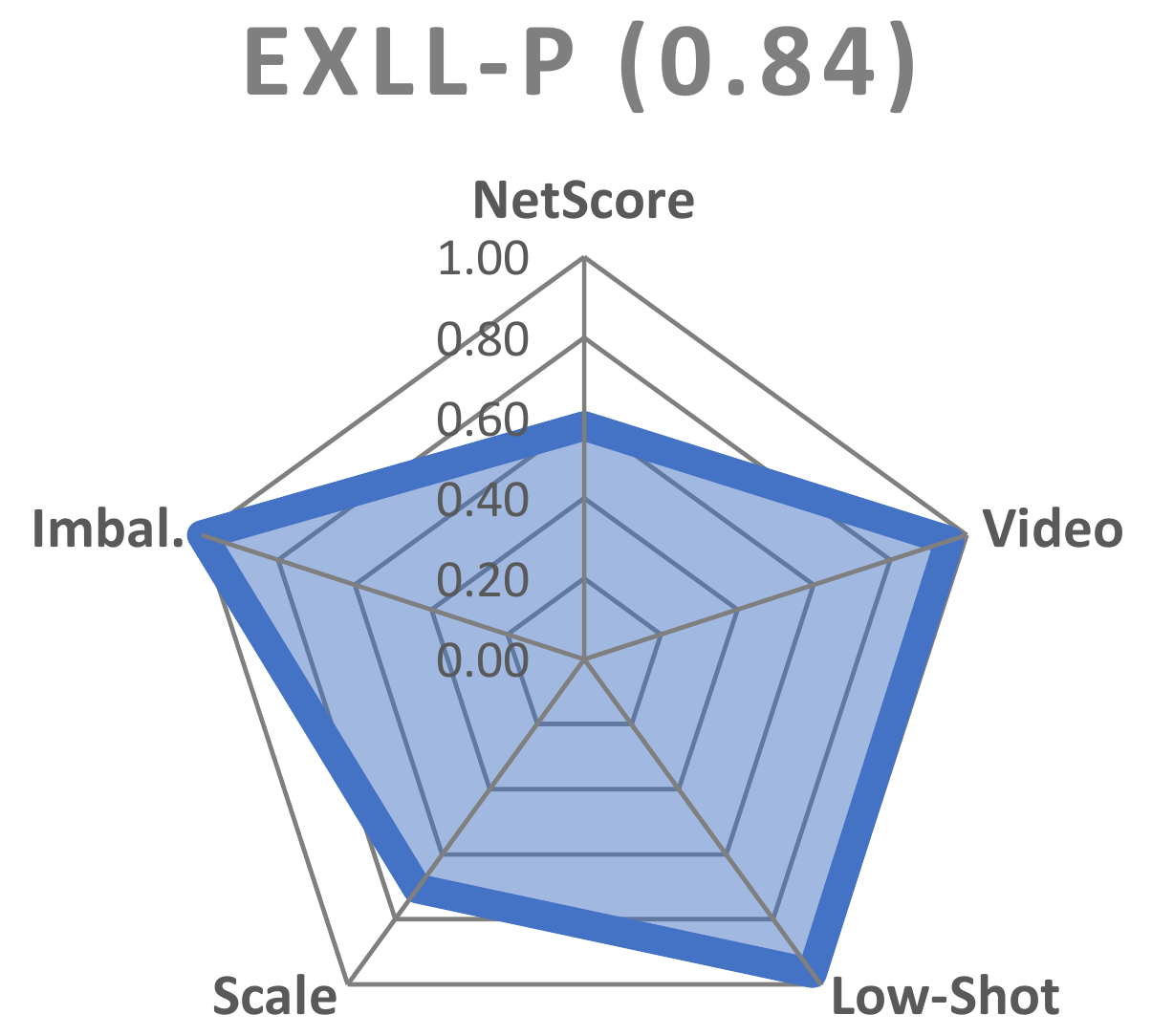} 
    \includegraphics[width=0.24\linewidth]{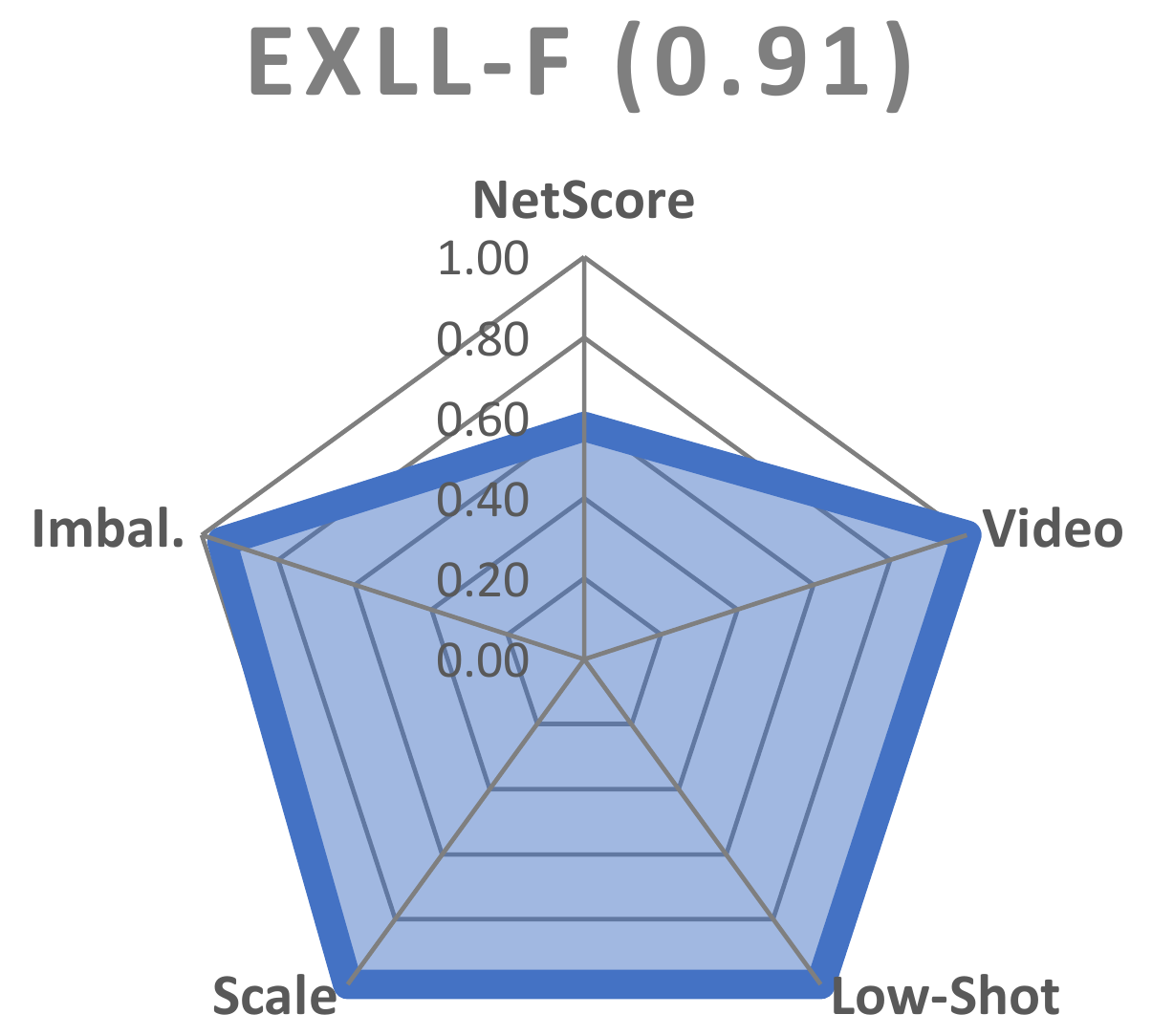} 
    \hfill
    
    \caption{The normalized performance metrics of online continual learners for accuracy and memory and computation requirements (\textbf{NetScore}); learning from temporally correlated videos (\textbf{Video}); generalizing from few data samples (\textbf{Low-Shot}); scalability (\textbf{Scale}); and learning from imbalanced data distributions (\textbf{Imbal.}). The learner's average performance across all metrics is shown at the top of each plot. \textbf{Higher} values are better.}
    \label{fig_results_spiderplot}
\end{figure*}

\subsection{Overall Results}

Spider plots were generated to visualize the performance metrics of online continual learners in terms of several factors: (\textbf{NetScore}), an index representing the learner's accuracy and memory and runtime requirements; (\textbf{Video}), the learner's ability to learn from sequential images or videos, evaluated based on its performance for the instance-ordered OpenLORIS dataset; (\textbf{Low-Shot}), the learner's ability to learn from a very small set of training inputs, evaluated using low-shot instance-ordered OpenLORIS; (\textbf{Scale}), its scalability to large-scale data, evaluated from Places-365; and (\textbf{Imbal.}), the learner's performance on imbalanced datasets, evaluated using Places-LT. To construct the plots, the performance metrics of learners were averaged for all backbone architectures and then normalized by assigning 0 to the worst score and 1 to the best score.

Figure \ref{fig_results_spiderplot} illustrates the generated spider plots. The online continual learner's name is presented at the top of each plot along with the averaged score for all five metrics. ExLL-F (0.91) showed the best overall performance. Replay 20pc (0.88) and SLDA (0.88) outperformed the second-best ExLL model, ExLL-P (0.84). The worst-performing ExLL model, ExLL-M (0.78) is also outperformed by NCM (0.81). 

The ExLL models performed poorly due to having low NetScores despite having better scores in the other four metrics. While sharing some similarities with SLDA, ExLL is less efficient with respect to computation and memory requirements. In addition to class vector means, ExLL models store prototype vector means as well as records of training samples to facilitate post hoc explainability during inference. 

% The difference in NetScores between the three ExLL methods is negligible.

\section{Conclusion}

% Restate your thesis statement: Begin by restating your thesis statement, which is the main argument or focus of your research paper. This helps to remind your reader what your paper is about and to ensure that your conclusion aligns with your thesis.
We propose an explainable neural network architecture suitable for online and continual learning applications on embedded devices. The Explainable Lifelong Learning (ExLL) model is a prototype-based classifier inspired by SLDA and is robust against catastrophic forgetting and mitigates the stability-plasticity dilemma. ExLL was designed to facilitate single-pass learning from a continuous data stream. The design of the architecture also makes it easy to generate IF-THEN rules and justify the classifier decisions with highly interpretable explanations. A collective inference strategy was implemented to combine the global MegaCloud inference with the local prototype-based inference using glocal pairwise decision fusion to enhance predictive accuracy. 

The classifier's performance was benchmarked against state-of-the-art online learning models using several different CNN backbones, object recognition datasets, and evaluation metrics. In terms of video classification accuracy, low-shot learning, scalability, and imbalanced data learning, ExLL outperformed other online learning models in nearly every scenario. However, in terms of metrics to quantify the model's storage and computational requirements, ExLL did not rank as high as Replay 20pc, SLDA, and NCM. One factor is due to these methods maintaining only one class mean per class while ExLL maintains a small topology of centroids per class as well as additional memory storage to facilitate explainability. Overall, ExLL showed state-of-the-art classification accuracy in continual learning scenarios. As for suitability for embedded applications, ExLL outperformed Perceptron, Fine-Tune, and Naive Bayes, but was ranked below SOvR, NCM, Replay, and SLDA. This is one of the trade-offs between number of parameters and experiment runtime requirements, and the need for a prototype-based architecture for explainability. 

% Summarize your main points: Next, summarize the key points you made in your paper. Briefly discuss the main findings of your research and the evidence that supports your arguments. This will help to tie together the various threads of your paper and provide a clear and concise summary of your research.

% Discuss the implications of your research: Once you have summarized your main points, it's important to discuss the implications of your research. What are the broader implications of your findings? How do they contribute to the existing body of knowledge in your field? This can help to show the significance of your research and its potential impact.

There are several strategies that can be considered to improve ExLL's efficiency. Pruning strategies may help identify low-utility prototypes that can be pruned without significant catastrophic forgetting, thus reducing the number of parameters requirement of the model \cite{liew2019effect, wiwatcharakoses2020soinn, logacjov2021learning}. Other explainability techniques can be applied to enhance interpretability, including the use of gradient class activation maps to visualize discriminative image features \cite{zhou2016learning} \cite{gee2019explaining}. Combined with selective feature weighing to ignore redundant features \cite{kenny2021explaining}, this may help reduce the dimensionality and computation required by the model. 

% Make recommendations for further research: Finally, you can make recommendations for future research in your field. Are there any unanswered questions or areas that require further investigation? This can help to show that your research is part of an ongoing conversation in your field and that there is more work to be done.

In conclusion, our research has shown that the proposed ExLL model achieved a very good performance when tested under diverse continual learning scenarios, even when compared against state-of-the-art continual learning models. Introducing the ability to explain and justify the model predictions is a necessary and important contribution for all online continual learning algorithms and that we have shown the merits of doing so.

\section*{Acknowledgments}
The authors acknowledge the support from the German Research Foundation (Deutsche Forschungsgemeinschaft/DFG) under project CML (TRR 169), the TRAnsparent, InterpretabLe Robots (TRAIL) EU project, and from the BMWK under project VeriKAS.

%Bibliography
\bibliographystyle{unsrt}  
\bibliography{references}  

\newpage

\appendix

\section*{Supplementary Materials}

\subsection{Visualization of the Places-365 topology}

\begin{figure}[h]
    \centering
    \includegraphics[width=0.45\columnwidth]{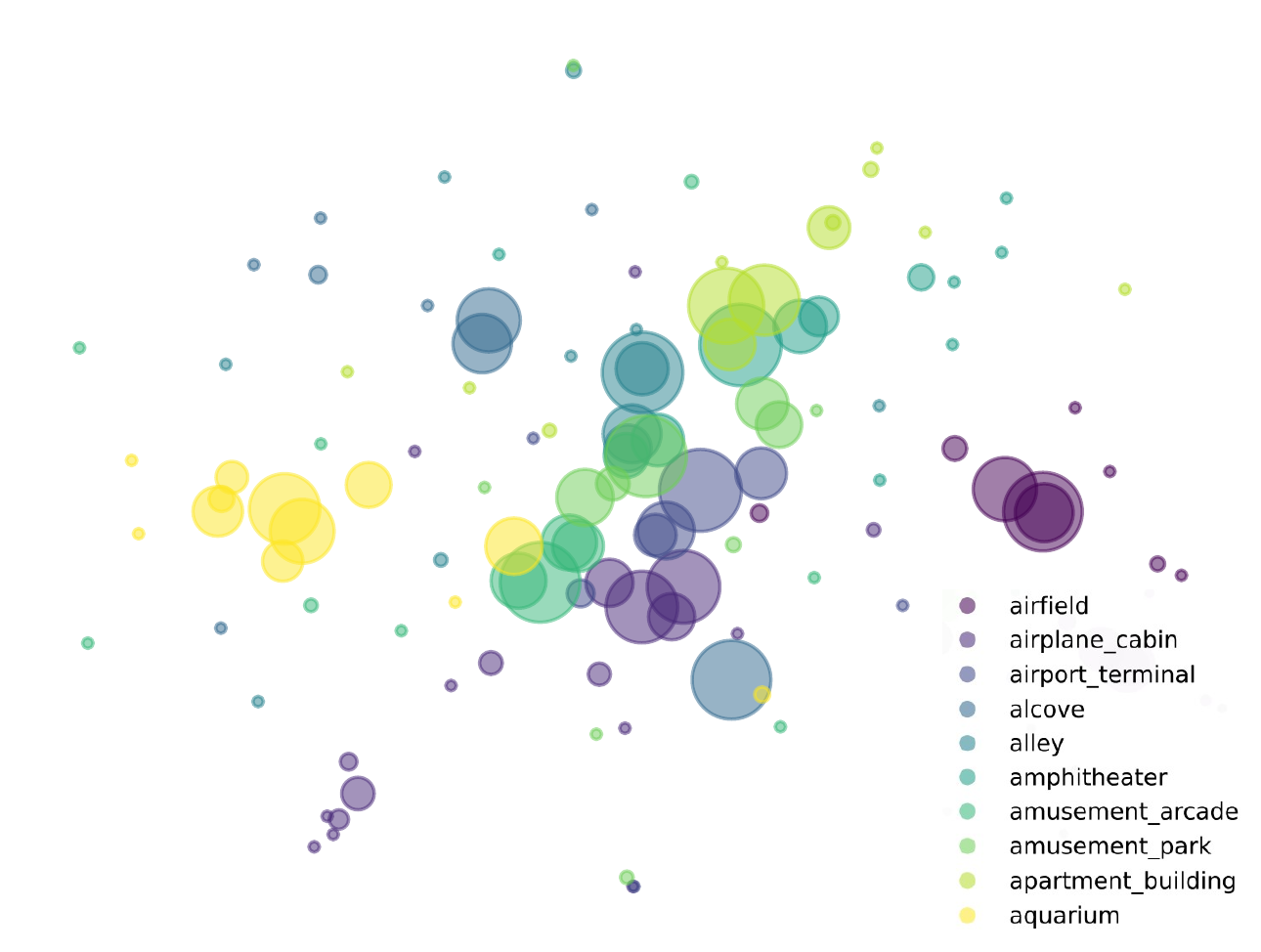}
    \includegraphics[width=0.45\columnwidth]{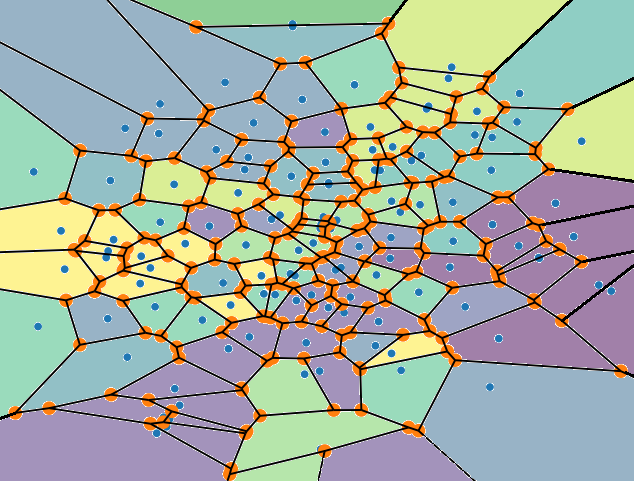}

    % \begin{subfigure}[b]{\columnwidth}
    %     \centering
    %     \includegraphics[width=0.45\columnwidth]{figures/fig_places_tsne10.pdf}
    %     \caption{}
    % \end{subfigure}
    % \hfill

    % \begin{subfigure}[b]{\columnwidth}
    %     \centering
    %     \includegraphics[width=0.45\columnwidth]{figures/fig_places_voronoi10.png}
    %     \caption{}
    % \end{subfigure}
    % \hfill
    
    \caption{Visualizing the topology of learned prototypes acquired from the Places-365 dataset, for the first 10 classes. On the left is a 2-dimensional representation of the identified prototypes generated using t-SNE. Each circle represents one prototype and the radius of each circle denotes the support for that prototype. On the right is the generated Voronoi tesselation.}
    \label{fig_visualize_places}
\end{figure}

Figure \ref{fig_visualize_places} visualizes the topology of the prototypes encoded using ExLL-F. In the left sub-figure, the high-dimensional centroids are transformed into 2-dimensional scatter plots using t-distributed stochastic neighbor embedding (t-SNE) \cite{van2008visualizing}. Each circle denotes one prototype and the radius of the circle indicates the support or size of the prototype's data cloud. There are multiple instances where neighbouring prototypes displayed overlapping areas of influences. The presence of overlap suggests that some prototypes are highly similar to each other even after self-organization. It is possible to reduce redundancies by merging the overlapping prototypes, for example, by using hierarchical clustering \cite{wang2012hierarchical} or divide-and-merge \cite{rehman2022divide}. In the right sub-figure, the same 2-dimensional prototype topology is represented using a Voronoi tesselation graph. The merged prototypes, for instance, may be visualized by removing the edges between neighbouring partitions with the same colour (i.e., same class labels). 

\newpage

\subsection{Visualization of the F-SIOL-310 topology}

\begin{figure}[h]
    \centering
    \includegraphics[width=0.45\columnwidth]{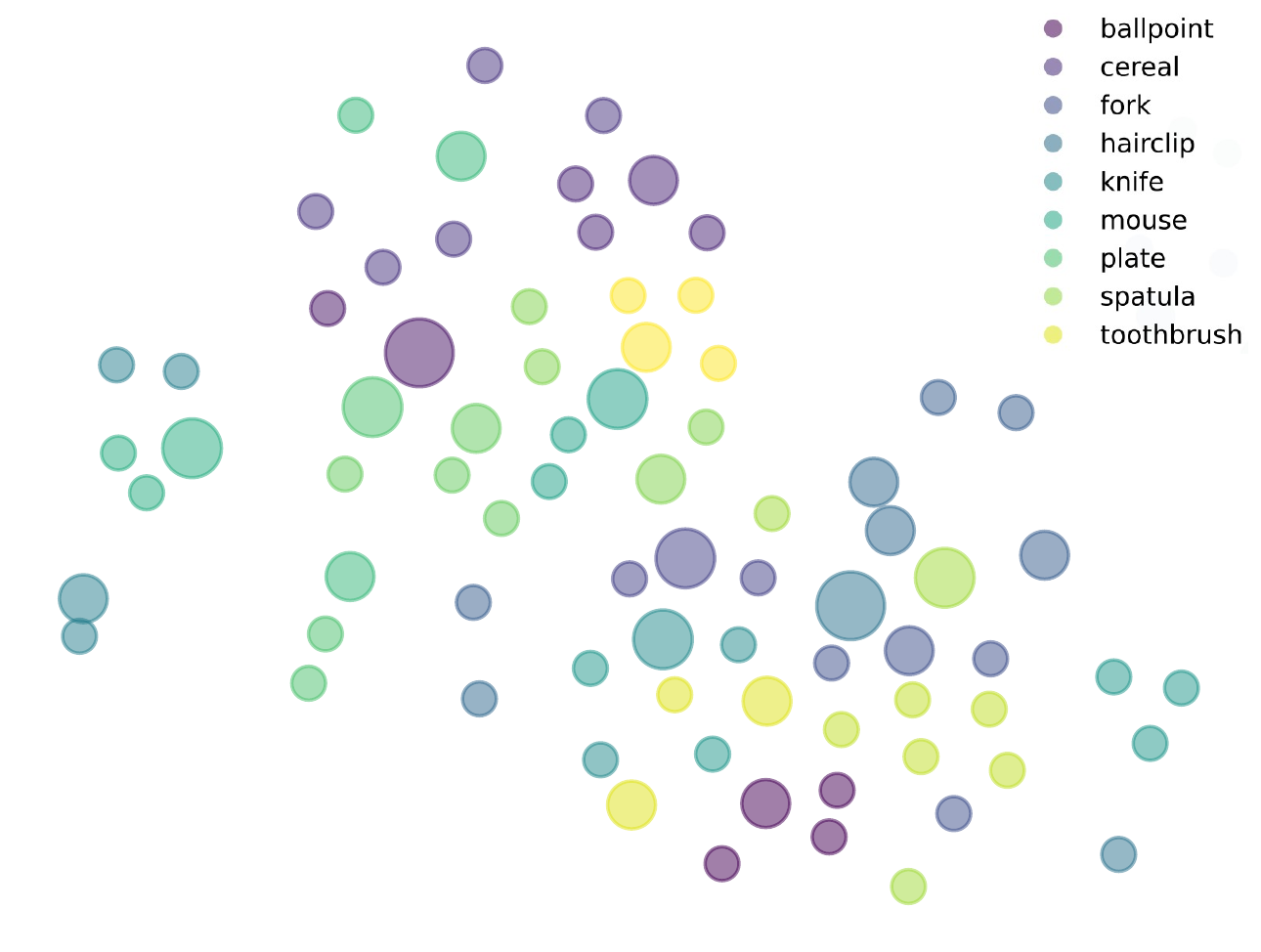}
    \includegraphics[width=0.45\columnwidth]{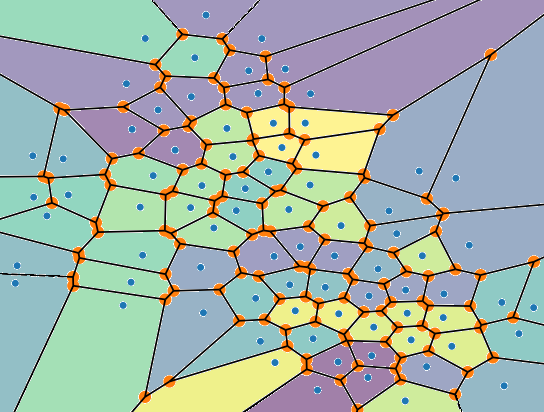}
    
    % \begin{subfigure}[b]{\columnwidth}
    %     \centering
    %     \includegraphics[width=0.8\columnwidth]{figures/fig_fsiol_tsne22.pdf}
    %     \caption{}
    % \end{subfigure}
    
    % \hfill
    % \hspace{\fill}
    % \begin{subfigure}[b]{\columnwidth}
    %     \centering
    %     \includegraphics[width=0.7\columnwidth]{figures/fig_fsiol_voronoi22.png}
    %     \caption{}
    % \end{subfigure}
    
    % \hspace{\fill}
    
    \caption{Visualizing the topology of learned prototypes acquired from the F-SIOL-310 dataset. On the left is a 2-dimensional representation of the identified prototypes generated using t-SNE. Each circle represents one prototype and the radius of each circle denotes the support for that prototype. On the right is the generated Voronoi tesselation.}
    \label{fig_visualize_fsiol}
\end{figure}

Figure \ref{fig_visualize_fsiol} visualizes the topology of the prototypes encoded using ExLL-F. In the left sub-figure, the high-dimensional centroids are transformed into 2-dimensional scatter plots using t-SNE. Each circle denotes one prototype and the radius of the circle indicates the support or size of the prototype's data cloud. The prototypes encoding the F-SIOL-310 dataset form a highly-separable topology with clear delineation between prototypes. This is considerably different compared to the Places topology in Figure \ref{fig_visualize_places}. The difference may be due to the visually distinctive objects in the F-SIOL-310 dataset while the Places dataset has significantly more variety of images within-class causing the overlap. Similarly in the right sub-figure, the Voronoi tesselation graph for F-SIOL-310 displayed a distribution of evenly spaced cells highlighting the prototype separability. 

\end{document}